\documentclass[letterpaper]{article} 
\usepackage{aaai2027}
\usepackage[hyphens]{url}  
\usepackage{graphicx} 
\usepackage{natbib}  
\usepackage{caption} 
\usepackage{algorithm}
\usepackage{algorithmic}

\usepackage{newfloat}
\usepackage{listings}
\DeclareCaptionStyle{ruled}{labelfont=normalfont,labelsep=colon,strut=off} 
\floatstyle{ruled}
\newfloat{listing}{tb}{lst}{}
\floatname{listing}{Listing}

\usepackage{booktabs}

\usepackage{amsthm}
\theoremstyle{definition} 
\newtheorem{definition}{Definition}

\usepackage{amssymb}
\usepackage{amsmath}

\usepackage{subcaption}
\usepackage{dsfont}
\usepackage{enumitem}

\usepackage{multirow}
\newcommand{\method}{BALIGN}

\title{Preference Data Selection for Mitigating the Alignment Tax in \\ Large Language Models}
\author{
    Written by AAAI Press Staff\textsuperscript{\rm 1}\thanks{With help from the AAAI Publications Committee.}\\
    AAAI Style Contributions by Peter Patel Schneider,
    Sunil Issar,\\
    J. Scott Penberthy,
    George Ferguson,
    Hans Guesgen,
    Francisco Cruz\equalcontrib\corresponding,
    Marc Pujol-Gonzalez\equalcontrib\corresponding
}
\affiliations{
    \textsuperscript{\rm 1}Association for the Advancement of Artificial Intelligence\\

    1101 Pennsylvania Ave, NW Suite 300\\
    Washington, DC 20004 USA\\
    proceedings-questions@aaai.org
}

\title{Preference Data Selection for Mitigating the Alignment Tax in \\ Large Language Models}
\author {
    Minsu Kim\textsuperscript{\rm 1,\rm 2}\thanks{Work done during an internship at Microsoft Research Asia.},
    Jianxun Lian\textsuperscript{\rm 2}\thanks{Corresponding authors.},
    Xing Xie\textsuperscript{\rm 2},
    Steven Euijong Whang\textsuperscript{\rm 1}\footnotemark[2]
}
\affiliations {
    \textsuperscript{\rm 1}KAIST\\
    \textsuperscript{\rm 2}Microsoft Research Asia\\
}

\nocopyright

\begin{document}

\maketitle

\begin{abstract}
Aligning large language models to human preferences is crucial for real-world deployment but frequently incurs an alignment tax, leading to the catastrophic forgetting of pre-trained general capabilities. While previous works primarily frame this problem as an optimization or architectural challenge, the inherent characteristics of preference data that drive this degradation remain largely underexplored. In this paper, we propose BALIGN, a balanced data selection strategy that explicitly mitigates catastrophic forgetting while optimizing alignment efficacy. Through theoretical and empirical analyses of the preference optimization gradient, we identify three key data-centric features that dictate parameter drift: the reference model's log-probability margin, the token length difference between chosen and rejected responses, and the TF--IDF similarity to general capability corpora. By aggregating these orthogonal features into a unified composite risk score, BALIGN systematically filters out high-risk preference samples that disrupt intrinsic model parameters or provide minimal alignment utility. Extensive experiments on standard human preference datasets demonstrate that BALIGN strongly preserves foundational capabilities without compromising alignment gains, consistently achieving the optimal Pareto frontier with minimal computational overhead.
\end{abstract}


\section{Introduction}

\begin{figure*}[t]
    \begin{minipage}[c]{0.304\textwidth}
        \includegraphics[width=\linewidth]{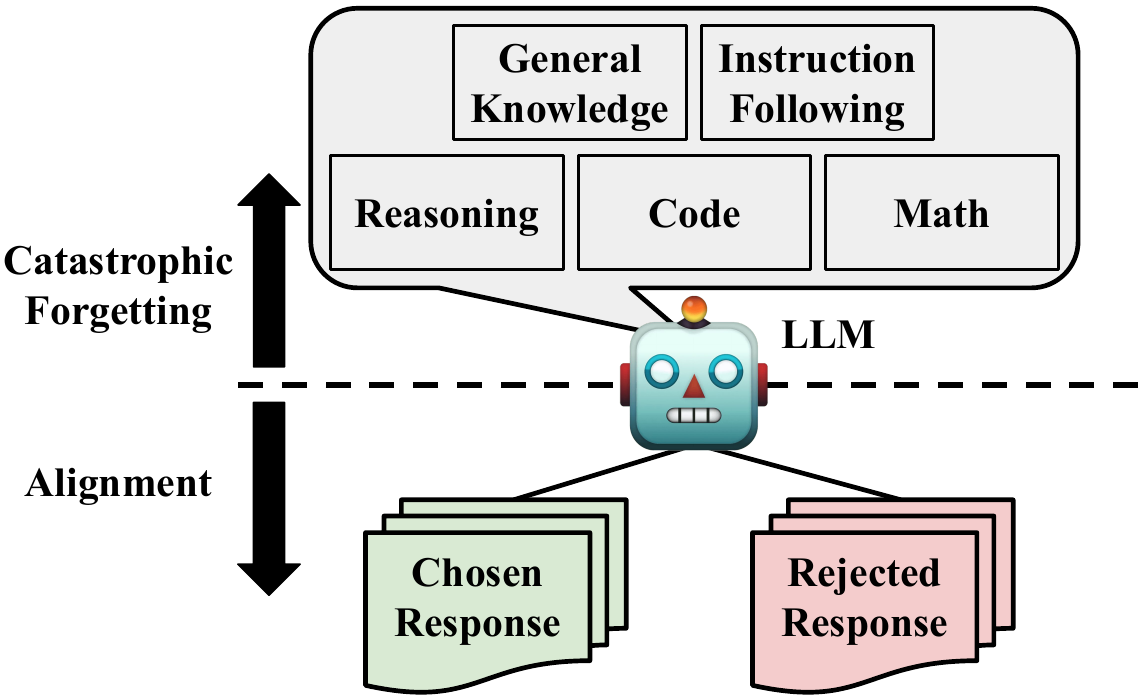}
    \end{minipage}\hfill
    \begin{minipage}[c]{0.696\textwidth}
        \includegraphics[width=\linewidth]{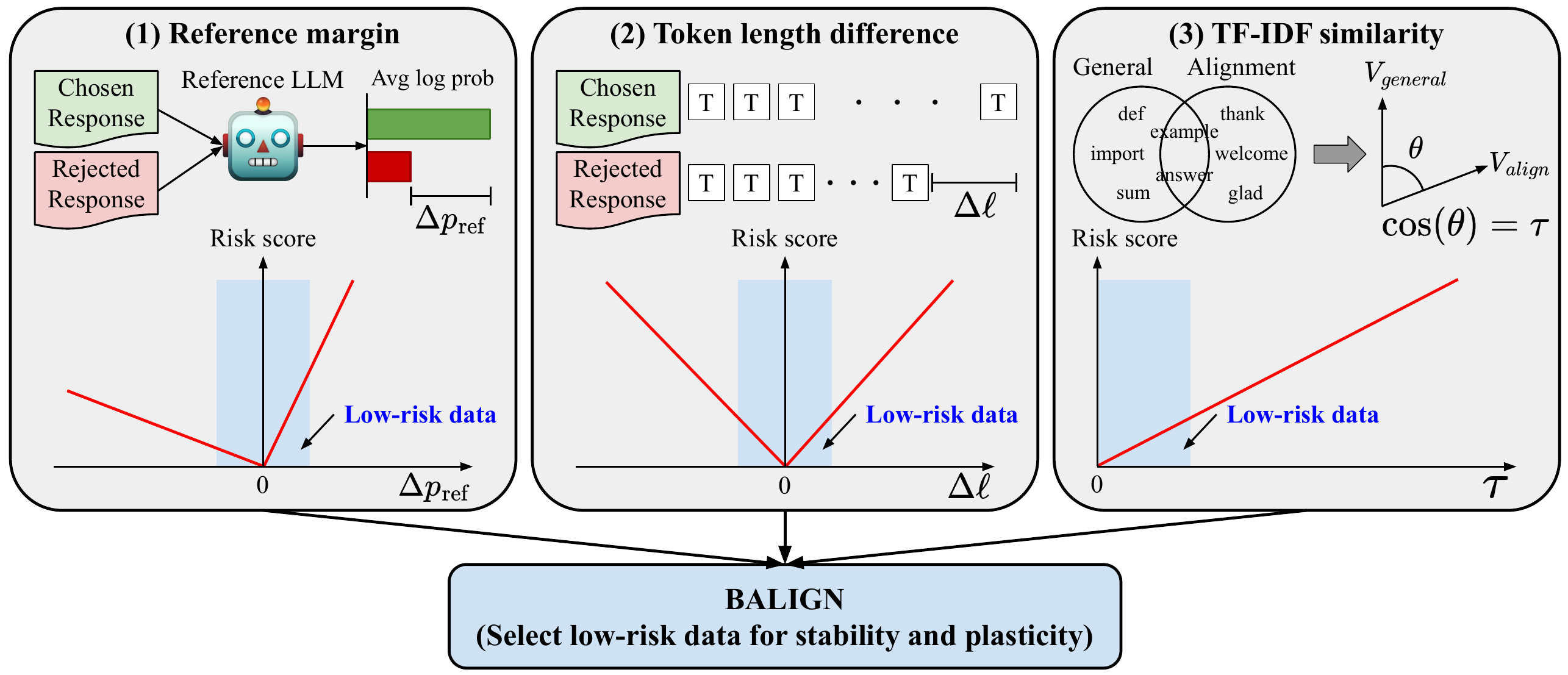}
    \end{minipage}
    \begin{subfigure}[t]{0.304\textwidth}
        \caption{}
        \label{fig:problem}
    \end{subfigure}\hfill
    \begin{subfigure}[t]{0.696\textwidth}
        \caption{}
        \label{fig:overview}
    \end{subfigure}
    \vspace{-0.2cm}
    \caption{(a) Catastrophic forgetting of general capabilities (e.g., general knowledge, instruction following, reasoning, code, and math) in LLM alignment. (b) Overview of \method{}, which selects low-risk data for stability and plasticity using risk scores based on three features: reference margin, token length difference, and TF--IDF similarity.}
    \label{fig:main_figure}
    \vspace{-0.3cm}
\end{figure*}

Large language models (LLMs) pre-trained on massive corpora have become foundational building blocks for a wide range of natural language processing applications\,\cite{DBLP:journals/corr/abs-2307-10169}. A common practice in deploying these models is to take pre-trained LLMs as base models and fine-tune them to enhance their capabilities, such as mathematical reasoning\,\cite{DBLP:journals/corr/abs-2402-03300}, clinical reasoning\,\cite{DBLP:journals/corr/abs-2412-18925}, or code generation\,\cite{DBLP:journals/corr/abs-2409-12186}. Among such fine-tuning objectives, alignment with human preferences is one of the most critical, as it determines whether the resulting model produces safe, helpful, and human-aligned responses\,\cite{DBLP:journals/corr/abs-2309-15025, DBLP:journals/corr/abs-2308-05374, DBLP:journals/corr/abs-2310-19852}.

While alignment improves the quality of model responses on preference-related tasks, it is known to induce a phenomenon referred to as the alignment tax\,\cite{DBLP:conf/nips/Ouyang0JAWMZASR22}, in which the model loses a portion of its previously acquired general capabilities.
This degradation constitutes a form of catastrophic forgetting\,\cite{DBLP:journals/corr/KirkpatrickPRVD16} as shown in Figure~\ref{fig:problem}. Such a decline in foundational skills limits the broader utility of LLMs, as their real-world application relies on these general capabilities. To mitigate this issue, effective alignment necessitates maintaining a balance between preserving general capabilities (i.e., stability) and integrating human-aligned preferences (i.e., plasticity). Despite the importance of this dual objective, most existing works on LLM fine-tuning have solely focused on adapting to new tasks for domain specialization\,\cite{DBLP:conf/nips/XieS0L23, DBLP:conf/icml/XiaMGA024, DBLP:conf/nips/LiuKR24}.

Prior work widely acknowledges that the underlying cause of catastrophic forgetting is not simply a deficit in memory capacity, but rather a disruption of intrinsic model parameters\,\cite{DBLP:journals/corr/KirkpatrickPRVD16, DBLP:journals/corr/abs-2308-08747}. Based on these findings, several approaches have been proposed to mitigate catastrophic forgetting in LLM fine-tuning\,\cite{DBLP:conf/iclr/KothaSR24, DBLP:conf/emnlp/Li0FT24, DBLP:conf/aaai/XiongX26}. 
However, these methods predominantly frame catastrophic forgetting as an optimization or architectural challenge, leaving the relationship between the characteristics of new task data and catastrophic forgetting largely underexplored.

Motivated by the intuition that the fundamental difference between general and new tasks disturbs intrinsic model parameters, we focus on the data itself to analyze and mitigate catastrophic forgetting. 
Given that LLMs solidify the majority of their general capabilities during the SFT phase, it is crucial to analyze the interplay between the knowledge embedded in the previously learned SFT data and the learning signals provided by the newly introduced preference data from a stability-plasticity perspective. To the best of our knowledge, no existing data-centric method simultaneously optimizes both stability and plasticity in LLM alignment.

We identify three key features that characterize the impact of training samples on the stability and plasticity of LLM alignment, as shown in Figure~\ref{fig:overview}. Specifically, the length-normalized log-probability margin under the base reference model (i.e., the initial pre-trained model prior to alignment) and the token length difference between the chosen and rejected responses serve as indicators for stability, whereas the cosine similarity between the TF--IDF representations of training samples and those of a general benchmark corpus acts as a proxy for plasticity. Our theoretical and empirical analyses reveal that selecting training samples with minimal length-normalized log-probability margins and small token length differences effectively mitigates catastrophic forgetting. 
In addition, selecting training samples that are highly distinct from the general SFT corpus based on their TF--IDF representations yields better model alignment.

Based on our findings, we propose \method{}, a balanced data selection strategy for effective LLM alignment that jointly addresses stability and plasticity. We first define a risk score tailored to each of the three features, namely \textit{reference margin}, \textit{token length difference}, and \textit{TF--IDF similarity}, such that training on samples with high risk scores leads to degradation in either stability or plasticity. We then formulate a composite risk score that accounts for stability and plasticity by aggregating the three normalized risk scores. Based on the composite risk scores computed across all samples, we filter out high-risk samples and select a predefined proportion of low-risk samples to form the training set.
Experiments on human preference datasets demonstrate that \method{} effectively mitigates catastrophic forgetting across diverse capabilities while preserving alignment performance with a filtered data subset.
We believe that our data selection strategy serves as a simple yet effective method for preprocessing training data in LLM alignment.


\textbf{Summary of Contributions:} (1) We identify three orthogonal data-centric features, namely reference margin, token length difference, and TF--IDF similarity, and analyze their impact on the stability-plasticity trade-off during LLM alignment; (2) We propose a balanced data selection strategy that mitigates catastrophic forgetting by filtering out high-risk preference data using a composite risk score tailored to these three features; (3) We empirically demonstrate that our approach successfully preserves general capabilities across diverse domains while maintaining alignment performance.

\vspace{-0.04cm}
\section{Related Work}

\paragraph{Large Language Model Alignment.}

LLM alignment refers to the process of adjusting a pre-trained language model so that its outputs conform to human preferences regarding helpfulness, harmlessness, honesty, and safety\,\cite{DBLP:journals/corr/abs-2309-15025, DBLP:journals/corr/abs-2507-19672}. A foundational approach is Reinforcement Learning from Human Feedback (RLHF)\,\cite{DBLP:conf/nips/ChristianoLBMLA17, DBLP:journals/corr/abs-2009-01325}, which combines supervised fine-tuning with a learned reward model and Proximal Policy Optimization (PPO)\,\cite{DBLP:journals/corr/SchulmanWDRK17} to align the policy with human preferences. While recent advancements such as Group Relative Policy Optimization (GRPO)\,\cite{DBLP:journals/corr/abs-2402-03300} mitigate the substantial memory overhead of the critic model in PPO, on-policy optimization remains engineering-heavy and computationally expensive. To remove the engineering complexity of explicit reward modeling and on-policy optimization altogether, Direct Preference Optimization (DPO)\,\cite{DBLP:conf/nips/RafailovSMMEF23} reparameterizes the alignment objective as a single logistic loss over preference pairs, treating the policy itself as an implicit reward function. 
Given its simplicity and status as the standard in offline alignment\,\cite{DBLP:journals/corr/abs-2407-21783, DBLP:journals/corr/abs-2412-15115}, we build our alignment framework upon DPO.

\paragraph{Catastrophic Forgetting of Large Language Models.}

A well-known challenge in LLM fine-tuning is catastrophic forgetting, in which a model forgets previously acquired knowledge while learning new tasks\,\cite{DBLP:journals/corr/KirkpatrickPRVD16, DBLP:conf/naacl/LiCCHLXGL22}. In LLM alignment, optimizing models for human preferences often leads to the degradation of pre-trained capabilities, a phenomenon referred to as the alignment tax\,\cite{DBLP:conf/nips/Ouyang0JAWMZASR22}. 
Recent studies attribute catastrophic forgetting to representational routing failures\,\cite{DBLP:conf/iclr/KothaSR24, DBLP:conf/iclr/JainKLDTRGK24}, convergence to sharp minima\,\cite{DBLP:conf/emnlp/Li0FT24, DBLP:journals/corr/abs-2605-02105}, and parameter-level weight interference\,\cite{DBLP:journals/corr/abs-2308-08747, DBLP:conf/emnlp/HuangCW25}.
However, the impact of fine-tuning data itself on catastrophic forgetting remains largely under-explored in the context of LLM alignment.

\paragraph{Mitigating Catastrophic Forgetting of Large Language Models.}

Conventional continual learning techniques, including experience replay and regularization-based approaches, were first applied to the scenario of LLM fine-tuning\,\cite{DBLP:conf/emnlp/ScialomCM22, DBLP:conf/acl/MokDLTYY23, lin2023speciality}. 
Recent studies have proposed methods specifically designed for LLMs by leveraging their internal properties. These include self-synthesized rehearsal\,\cite{DBLP:conf/acl/HuangCWYLSYS24}, self-distillation\,\cite{DBLP:conf/acl/YangPFWCZL24, DBLP:journals/corr/abs-2601-19897}, and model merging\,\cite{DBLP:conf/acl/XiaoLZX24, DBLP:conf/emnlp/LinL0DLZP00ZDPZ24, DBLP:conf/emnlp/AlexandrovRMZVT24}. While these approaches have substantially advanced the trade-off between stability and plasticity, they have primarily focused on supervised fine-tuning for task-specific performance. Consequently, the joint dynamics between catastrophic forgetting and alignment remain underexplored in the data-centric literature. Our work addresses this gap by formulating a sample-level composite risk score that explicitly accounts for both objectives.

\section{Problem Definition}

\subsection{Notation and Preliminaries}
Let $\mathcal{X}$ denote the space of prompts and $\mathcal{Y}$ the space of responses. Let $\pi_{\mathrm{ref}}: \mathcal{X} \rightarrow \Delta(\mathcal{Y})$ denote a pre-trained reference language model that serves as the starting point for alignment. Let $\mathcal{D} = \bigl\{ (x_i, y_w^{(i)}, y_l^{(i)}) \bigr\}_{i=1}^{N}$ be a human preference dataset, where $y_w^{(i)} \succ y_l^{(i)} \mid x_i$ denotes that the response $y_w^{(i)}$ is preferred over $y_l^{(i)}$ for a given prompt $x_i$. We refer to $y_w^{(i)}$ as the chosen response and $y_l^{(i)}$ as the rejected response. Using this preference dataset, we aim to optimize a parameterized policy $\pi_{\theta}: \mathcal{X} \rightarrow \Delta(\mathcal{Y})$, starting from the initial reference model $\pi_{\theta_0} = \pi_{\mathrm{ref}}$.

We adopt DPO\,\cite{DBLP:conf/nips/RafailovSMMEF23} as our primary alignment objective, which optimizes the policy $\pi_{\theta}$ to satisfy the observed preferences without explicitly learning a reward model. The DPO loss is given by
\begin{equation}
\mathcal{L}_{\mathrm{DPO}}(\pi_{\theta}; \pi_{\mathrm{ref}}, \mathcal{D}) = - \mathbb{E}_{(x, y_w, y_l) \sim \mathcal{D}}\left[ \log \sigma\bigl( \beta m_{\theta}(x, y_w, y_l) \bigr) \right],
\label{eq:dpo}
\end{equation}
where $\sigma(\cdot)$ denotes the sigmoid function, $\beta > 0$ is a hyperparameter controlling the strength of the implicit Kullback--Leibler regularization toward $\pi_{\mathrm{ref}}$, and
\begin{equation}
m_{\theta}(x, y_w, y_l) = \log \frac{\pi_{\theta}(y_w \mid x)}{\pi_{\mathrm{ref}}(y_w \mid x)} - \log \frac{\pi_{\theta}(y_l \mid x)}{\pi_{\mathrm{ref}}(y_l \mid x)}
\label{eq:margin}
\end{equation}
is the reward margin between chosen and rejected responses.

\subsection{Catastrophic Forgetting and Alignment}
We characterize the policy $\pi_{\theta}$ along two orthogonal axes: \emph{general capability}, which tracks the retention of previously acquired general knowledge, and \emph{alignment quality}, which measures adherence to human preferences.

\paragraph{General Capability.}
Let $\mathcal{E} = \{e_1, e_2, \ldots, e_K\}$ denote a suite of general-capability benchmarks that are disjoint from the alignment distribution. For each benchmark $e_k$, let $c_{e_k}(\pi) \in [0, 1]$ denote the normalized performance of policy $\pi$ on $e_k$. We define the \emph{aggregate general capability} as: 
\begin{equation}
C(\pi) = \frac{1}{K} \sum_{k=1}^{K} c_{e_k}(\pi).
\label{eq:general}
\end{equation}
Based on this, we define the \emph{catastrophic forgetting} of general capabilities induced by alignment as:
\begin{equation}
F(\pi_{\theta}) = C(\pi_{\mathrm{ref}}) - C(\pi_{\theta}).
\label{eq:forgetting}
\end{equation}
A positive $F(\pi_{\theta})$ indicates that alignment has degraded the model's general capability relative to the reference.

\paragraph{Alignment Quality.}
Let $\mathcal{D}_{\mathrm{test}} = \{(x_j, y_w^{(j)}, y_l^{(j)})\}_{j=1}^{M}$ denote a held-out preference test set. We measure \emph{alignment quality} by the preference accuracy of the policy, defined as the proportion of samples where the policy assigns a higher log-probability to the chosen response $y_w^{(j)}$ than to the rejected response $y_l^{(j)}$, given by: 
\begin{equation}
A(\pi_{\theta}) = \frac{1}{M} \sum_{j=1}^{M} \mathds{1}\left[ \log \pi_{\theta}\bigl(y_w^{(j)} \mid x_j\bigr) > \log \pi_{\theta}\bigl(y_l^{(j)} \mid x_j\bigr) \right].
\label{eq:alignment}
\end{equation}
Building on this, we define the \emph{alignment gain} to quantify the net improvement in alignment quality as:
\begin{equation}
R(\pi_{\theta}) = A(\pi_{\theta}) - A(\pi_{\mathrm{ref}}).
\label{eq:alignment_gain}
\end{equation}

\subsection{Balanced Data Selection for Alignment}
We frame preference data selection as a bi-level optimization problem to navigate the trade-off between alignment and catastrophic forgetting. Given a sampling ratio $\rho \in (0, 1]$, we aim to identify an optimal subset $\mathcal{S} \subseteq \mathcal{D}$ of size $\lceil \rho N \rceil$ that maximizes the joint objective:
\begin{equation}
\max_{\mathcal{S} \subseteq \mathcal{D}, \, |\mathcal{S}| = \lceil \rho N \rceil} \Bigl( R(\pi_{\theta}^{\mathcal{S}}) - F(\pi_{\theta}^{\mathcal{S}}) \Bigr),
\label{eq:biobjective}
\end{equation}
where the inner objective defines the policy $\pi_{\theta}^{\mathcal{S}}$, which is optimized on the selected subset $\mathcal{S}$ via the DPO loss:
\begin{equation}
\pi_{\theta}^{\mathcal{S}} = \arg\min_{\theta \in \Theta} \mathcal{L}_{\mathrm{DPO}}(\pi_{\theta}; \pi_{\mathrm{ref}}, \mathcal{S}).
\end{equation}

\section{Method}

We propose \method{}, a balanced data selection strategy for LLM alignment that jointly addresses stability and plasticity. Specifically, we derive three key features that capture how preference data influences the trade-off between mitigating catastrophic forgetting and enhancing alignment quality.

\subsection{Theoretical Analysis}

We begin by examining the gradient of the DPO loss with respect to the policy parameters $\theta$, which governs both the speed of alignment and the magnitude of parameter drift away from the reference $\pi_{\mathrm{ref}}$. Differentiating Eq.~(\ref{eq:dpo}) yields
\begin{equation}
\begin{split}
\nabla_{\theta} \mathcal{L}_{\mathrm{DPO}} &= -\beta \mathbb{E}_{(x, y_w, y_l) \sim \mathcal{D}} \biggl[ \underbrace{\sigma\bigl(-\beta m_{\theta}(x, y_w, y_l)\bigr)}_{\text{per-sample weight } w_{\theta}} \\
& \cdot \underbrace{\bigl( \nabla_{\theta} \log \pi_{\theta}(y_w \mid x) - \nabla_{\theta} \log \pi_{\theta}(y_l \mid x) \bigr)}_{\text{per-sample direction } \Delta_{\theta}} \biggr],
\end{split}
\label{eq:dpo_grad}
\end{equation}
where each preference sample contributes a scalar weight $w_{\theta}$ and a direction $\Delta_{\theta}$ to the overall parameter update. 
Based on this structural decomposition, the three features we propose below characterize the gradient direction $\Delta_{\theta}$ in terms of its informational utility, accumulation scale, and support structure in the parameter space.

\paragraph{\textbf{Reference Margin.}}
The first feature is the reference margin $\Delta p_{\mathrm{ref}}$, defined as the difference between the average per-token log-probabilities of the chosen and rejected responses under the reference model $\pi_{\mathrm{ref}}$:
\begin{equation}
\Delta p_{\mathrm{ref}}(x, y_w, y_l) = \frac{1}{|y_w|} \log \pi_{\mathrm{ref}}\bigl(y_w \mid x\bigr) - \frac{1}{|y_l|} \log \pi_{\mathrm{ref}}\bigl(y_l \mid x\bigr).
\label{eq:mean_margin_step1}
\end{equation}
The magnitude and sign of $\Delta p_{\mathrm{ref}}$ dictate the risk of catastrophic forgetting by reflecting how the reference model inherently perceives the preference pair. Samples with a large positive $\Delta p_{\mathrm{ref}}$ indicate that the reference model already intrinsically prefers $y_w$ over $y_l$. In such cases, updating the model further along their respective gradient directions fails to impart novel alignment signals and over-optimizes existing preferences, thereby amplifying representational drift with negligible learning benefit. Conversely, samples with a large negative $\Delta p_{\mathrm{ref}}$ represent cases where the reference model strongly favors $y_l$. Forcing the model to overturn this deeply encoded prior knowledge requires substantial parameter change, similarly driving catastrophic forgetting.

Motivated by these insights, we aim to penalize samples with a large $|\Delta p_{\mathrm{ref}}|$ during the data selection process. To achieve this, we leverage a pinball loss formulation to define a reference margin-based risk score as follows:
\begin{definition}
\label{def:margin_score}
Let $\Delta p_{\mathrm{ref}}^{(i)} = \Delta p_{\mathrm{ref}}(x_{i}, y_{w}^{(i)}, y_{l}^{(i)})$ denote the reference margin of sample $i$. The \emph{reference margin-based risk score} of sample $i$ is:
\begin{equation}
s_{\Delta p_{\mathrm{ref}}}(i) = \alpha\max\bigl(0,\, -\Delta p_{\mathrm{ref}}^{(i)}\bigr) \;+\; (1 - \alpha)\max\bigl(0,\, \Delta p_{\mathrm{ref}}^{(i)}\bigr),
\label{eq:margin_score}
\end{equation}
where $\alpha \in [0, 1]$ is a hyperparameter designed to capture the asymmetric risk associated with the sign of $\Delta p_{\mathrm{ref}}$.
\end{definition}

\paragraph{\textbf{Token Length Difference.}}
The second feature is the token length difference between the chosen and rejected responses, defined as $\Delta\ell(y_w, y_l) = |y_w| - |y_l|$, where $|y|$ denotes the number of tokens in response $y$. By decomposing the log-probabilities into per-token contributions, $\log \pi_{\theta}(y \mid x) = \sum_{t=1}^{|y|} \log \pi_{\theta}(y_{t} \mid x, y_{<t})$, we can rewrite the reward margin in Eq.~(\ref{eq:margin}) as
\begin{equation}
m_{\theta}(x, y_w, y_l) = |y_w| \overline{r}_{\theta}(y_w \mid x) - |y_l| \overline{r}_{\theta}(y_l \mid x),
\label{eq:margin_decomp}
\end{equation}
where $\overline{r}_{\theta}(y \mid x) = \frac{1}{|y|}\sum_{t=1}^{|y|} \log \frac{\pi_{\theta}(y_t \mid x, y_{<t})}{\pi_{\mathrm{ref}}(y_t \mid x, y_{<t})}$ is the per-token log-ratio averaged over the response. Letting $\overline{r}_w$ and $\overline{r}_l$ denote the per-token ratios, and defining $\overline{r} = \tfrac{1}{2}(\overline{r}_w + \overline{r}_l)$ and $\overline{\ell} = \tfrac{1}{2}(|y_w| + |y_l|)$, we can decompose the margin:
\begin{equation}
m_{\theta} = \overline{r} \cdot \Delta\ell + (\overline{r}_w - \overline{r}_l) \cdot \overline{\ell}.
\label{eq:margin_two_axis}
\end{equation}

Eq.~(\ref{eq:margin_two_axis}) shows that the margin decomposes into a length-driven term proportional to $\Delta\ell$ and a quality-driven term proportional to the per-token ratio gap $\overline{r}_w - \overline{r}_l$. Whenever $|\Delta\ell|$ is large, the margin signal becomes dominated by sheer response length rather than by genuine preference quality, and the resulting gradient direction $\Delta_{\theta}$ in Eq.~(\ref{eq:dpo_grad}) is correspondingly amplified by the total token count. As parameter drift scales with the cumulative gradient norm, large length asymmetry inflates the alignment update without conveying additional preference information, providing a theoretical pathway from $\Delta\ell$ to catastrophic forgetting. To account for this length-induced risk during data selection, we introduce the length difference-based risk score as follows:
\begin{definition}
\label{def:length_score}
Let $\Delta\ell_{i} = |y_{w}^{(i)}| - |y_{l}^{(i)}|$ denote the token-level length difference between the chosen and rejected responses of sample $i$. The \emph{length difference-based risk score} of sample $i$ is defined as:
\begin{equation}
s_{\Delta\ell}(i) = \gamma\max\bigl(0,\, -\Delta\ell_{i}\bigr) \;+\; (1 - \gamma)\max\bigl(0,\, \Delta\ell_{i}\bigr),
\label{eq:length_score}
\end{equation}
where $\gamma \in [0, 1]$ is a hyperparameter designed to capture the asymmetric risk associated with the sign of $\Delta\ell$.
\end{definition}

\paragraph{\textbf{TF--IDF Similarity.}}
The third feature is the TF--IDF similarity, which measures how closely the lexical distribution of each preference sample aligns with that of the general capability datasets we aim to preserve. Given a corpus $\mathcal{G}$ comprising a subset of general capability datasets, we fit a TF--IDF vectorizer on the corpus $\mathcal{D} \cup \mathcal{G}$ and project both preference and general data into the same vector space. To align the format between preference and general data, we compute the TF--IDF vector of each preference sample using its prompt and chosen response. Using these vector representations, the TF--IDF similarity of sample $i$ is defined as the average cosine similarity to the general capability corpus:
\begin{align}
\tau_i \;=\; \frac{1}{|\mathcal{G}|} \sum_{(x_j, y_j) \in \mathcal{G}} \frac{\bigl\langle \mathbf{v}(x_i, y_w^{(i)}),\; \mathbf{v}(x_j, y_j) \bigr\rangle}{\bigl\|\mathbf{v}(x_i, y_w^{(i)})\bigr\|\,\bigl\|\mathbf{v}(x_j, y_j)\bigr\|},
\label{eq:tfidf_sim}
\end{align}
where $\mathbf{v}(x_i, y_w^{(i)})$ and $\mathbf{v}(x_j, y_j)$ denote the vector representations of the preference sample and the general sample, respectively. We note that since all components of a TF--IDF vector are non-negative, the resulting cosine similarity $\tau_i$ is inherently non-negative and bounded within $[0, 1]$.

When the similarity $\tau_i$ is high, the preference and general samples share an extensive vocabulary, meaning that their gradients actively update overlapping parameter subspaces. However, when $\tau_i$ is low, the preference sample introduces distinct lexical structures, such as novel instructional formats or specific alignment markers, that are largely absent from $\mathcal{G}$. Then the gradients of the preference and general samples are supported on substantially disjoint parameter regions, rendering them nearly orthogonal. As these updates do not conflict, the model can fully absorb the alignment signals without structural resistance from pre-trained priors. This orthogonality ensures that the optimization step maximizes alignment efficacy while leaving the general capability intact. Since higher similarity corresponds to a greater risk of gradient interference, we define the similarity-based risk score as the similarity value itself:
\begin{definition}
\label{def:tfidf_score}
Let $\tau_i \in [0, 1]$ denote the TF--IDF similarity of sample $i$. The \emph{similarity-based risk score} of sample $i$ is
\begin{equation}
s_{\tau}(i) \;=\; \tau_i.
\label{eq:tfidf_score}
\end{equation}
\end{definition}


\subsection{Empirical Analysis}

To verify the theoretical analysis and derived scores, we conduct a bucket-based experiment in which each of the three features is used in isolation to partition the preference dataset into buckets. We use the HH-RLHF\,\cite{DBLP:journals/corr/abs-2204-05862} dataset comprising preference triples, and adopt Llama-3.1-8B-Instruct\,\cite{DBLP:journals/corr/abs-2407-21783} as the reference model. For each feature $f \in \{\Delta p_{\mathrm{ref}}, \Delta\ell, \tau\}$, we compute the corresponding risk score for every sample and then partition the dataset into five equal-sized quantile buckets $\{\mathcal{B}_{f}^{1}, \ldots, \mathcal{B}_{f}^{5}\}$, ordered from the lowest to the highest risk score. We set the hyperparameters $\alpha=0.25$ and $\gamma=0.5$ for the reference margin and length difference risk scores, respectively. These values reflect our empirical observations that the reference margin exhibits asymmetry where positive values incur greater risk, whereas the length difference demonstrates symmetry as positive and negative values offset each other.

Using each bucket, we train a separate policy and evaluate its alignment quality and the extent of catastrophic forgetting on a general benchmark suite, including MMLU\,\cite{DBLP:conf/iclr/HendrycksBBZMSS21}, IFEval\,\cite{DBLP:journals/corr/abs-2311-07911}, ARC-Challenge\,\cite{DBLP:journals/corr/abs-1803-05457}, HumanEval\,\cite{DBLP:journals/corr/abs-2107-03374}, and GSM8K\,\cite{DBLP:journals/corr/abs-2110-14168}. We report the catastrophic forgetting ($F$) and alignment gain ($R$) across five buckets for each risk score as shown in Table~\ref{tab:bucket_results}. The results show that our proposed risk scores capture complementary dimensions of model degradation. Specifically, $s_{\Delta p_{\mathrm{ref}}}$ and $s_{\Delta\ell}$ serve as primary indicators of catastrophic forgetting,
while $s_{\tau}$ captures the failure of preference learning. 

\begin{table}[t]
\small
\setlength{\tabcolsep}{5.9pt}
\centering
\centering
\begin{tabular}{l|cc|cc|cc}
\toprule
Bucket & \multicolumn{2}{c|}{$s_{\Delta p_{\mathrm{ref}}}$} & \multicolumn{2}{c|}{$s_{\Delta\ell}$} & \multicolumn{2}{c}{$s_{\tau}$} \\
\midrule
& F $\downarrow$ & R $\uparrow$ & F $\downarrow$ & R $\uparrow$ & F $\downarrow$ & R $\uparrow$ \\ 
\midrule
$\mathcal{B}^{1}$ (lowest risk) & 0.08 & 3.33 & 0.38 & 3.20 & 0.95 & 3.80 \\
$\mathcal{B}^{2}$ & 0.37 & 3.46 & 0.82 & 2.95 & 1.43 & 3.93 \\
$\mathcal{B}^{3}$ & 0.52 & 4.02 & 1.77 & 2.26 & 0.73 & 3.42 \\
$\mathcal{B}^{4}$ & 0.84 & 3.33 & 1.35 & 3.20 & 0.84 & 2.61 \\
$\mathcal{B}^{5}$ (highest risk) & 3.48 & 1.75 & 2.19 & 3.12 & 0.85 & 1.67 \\
\bottomrule
\end{tabular}
\vspace{-0.2cm}
\caption{Analysis of catastrophic forgetting $F$ (lower is better) and alignment gain $R$ (higher is better) for each risk score.}
\label{tab:bucket_results}
\vspace{-0.2cm}
\end{table}

\subsection{BALIGN Data Selection Strategy}

Motivated by the complementary roles of the three risk scores, we propose a unified data selection strategy, \method{}. To jointly mitigate catastrophic forgetting and maximize alignment gain, we place the individual risk scores on a common scale and aggregate them into a single composite metric, formally defined as follows:
\begin{definition}
\label{def:composite}
The \emph{composite risk score} of sample $i$ is defined as the weighted sum of the three normalized risk scores:
\begin{equation}
s(i) = \widetilde{s}_{\Delta p_{\mathrm{ref}}}(i) + \widetilde{s}_{\Delta\ell}(i) + \lambda \, \widetilde{s}_{\tau}(i),
\label{eq:composite}
\end{equation}
where $\widetilde{s}_{f}(i) \in [0, 1]$ is the min--max normalized risk score for each feature $f \in \{\Delta p_{\mathrm{ref}}, \Delta\ell, \tau\}$, and $\lambda$ is a hyperparameter that controls the balance between the risk scores.
\end{definition}
The composite score $s(i)$ thus integrates the dual aspects of stability and plasticity into a single scalar that quantifies the total expected risk of including sample $i$ in preference learning. Following a score-based data selection template to minimize risks, our selected subset under a budget ratio $\rho \in (0, 1]$ is given by
\begin{equation}
\mathcal{S} = \bigl\{ (x_{i}, y_{w}^{(i)}, y_{l}^{(i)}) \in \mathcal{D} : s(i) \leq q_{\rho}(s) \bigr\},
\label{eq:final_selection}
\end{equation}
where $q_{\rho}(s)$ is the $\rho$-quantile of the composite scores $\{s(i)\}_{i=1}^{N}$. The final aligned policy is obtained by solving
\begin{equation}
\pi_{\theta}^{\mathcal{S}} = \arg\min_{\theta \in \Theta} \mathcal{L}_{\mathrm{DPO}}\bigl(\pi_{\theta}; \pi_{\mathrm{ref}}, \mathcal{S}\bigr).
\label{eq:final_train}
\end{equation}

\begin{table*}[t]
  \small
  \setlength{\tabcolsep}{2.6pt}
  \centering
  \begin{tabular}{l|cc|cc|cc|cc}
    \toprule
    {Method} & \multicolumn{4}{c|}{\sf Helpfulness} & \multicolumn{4}{c}{\sf Harmlessness} \\
    \cmidrule{1-9}
    {} & \multicolumn{2}{c|}{\sf Llama-3.1-8B-Instruct} & \multicolumn{2}{c|}{\sf Qwen2.5-7B-Instruct} & \multicolumn{2}{c|}{\sf Llama-3.1-8B-Instruct} & \multicolumn{2}{c}{\sf Qwen2.5-7B-Instruct} \\
    \cmidrule{1-9}
    {} & {General (F$\downarrow$)} & {Align (R$\uparrow$)} & {General (F$\downarrow$)} & {Align (R$\uparrow$)} & {General (F$\downarrow$)} & {Align (R$\uparrow$)} & {General (F$\downarrow$)} & {Align (R$\uparrow$)} \\
    \midrule
    {Base} & {76.86 (--)} & {61.64 (--)} & {81.52 (--)} & {62.84 (--)} & {76.86 (--)} & {45.25 (--)} & {81.52 (--)} & {44.11 (--)} \\
    \cmidrule{1-9}
    {Full} & {71.51 (5.35$\downarrow$)} & {65.40 (3.76$\uparrow$)} & {79.21 (2.31$\downarrow$)} & {66.17 (3.33$\uparrow$)} & {71.44 (5.42$\downarrow$)} & {53.92 (8.67$\uparrow$)} & {80.03 (1.49$\downarrow$)} & {52.17 (8.06$\uparrow$)} \\
    {Random} & {74.77 (2.09$\downarrow$)} & {64.80 (3.16$\uparrow$)} & {80.39 (1.13$\downarrow$)} & {64.76 (1.92$\uparrow$)} & \underline{75.78 (1.08$\downarrow$)} & {52.25 (7.00$\uparrow$)} & {80.82 (0.70$\downarrow$)} & {47.57 (3.46$\uparrow$)} \\
    \cmidrule{1-9}
    {DSIR} & {74.89 (1.97$\downarrow$)} & {63.44 (1.80$\uparrow$)} & {80.30 (1.22$\downarrow$)} & {64.50 (1.66$\uparrow$)} & {75.10 (1.76$\downarrow$)} & {50.77 (5.52$\uparrow$)} & \underline{80.86 (0.66$\downarrow$)} & {47.83 (3.72$\uparrow$)} \\
    {LESS} & {72.18 (4.68$\downarrow$)} & {64.14 (2.50$\uparrow$)} & {79.84 (1.68$\downarrow$)} & {63.61 (0.77$\uparrow$)} & {74.84 (2.02$\downarrow$)} & {49.19 (3.94$\uparrow$)} & {79.93 (1.59$\downarrow$)} & {48.62 (4.51$\uparrow$)} \\
    {TSDS} & {75.60 (1.26$\downarrow$)} & {64.56 (2.92$\uparrow$)} & {80.42 (1.10$\downarrow$)} & {64.55 (1.71$\uparrow$)} & {75.09 (1.77$\downarrow$)} & {50.63 (5.38$\uparrow$)} & {80.76 (0.76$\downarrow$)} & {46.96 (2.85$\uparrow$)} \\
    {NICE} & {73.18 (3.68$\downarrow$)} & {65.05 (3.41$\uparrow$)} & {80.58 (0.94$\downarrow$)} & {64.08 (1.24$\uparrow$)} & {74.77 (2.09$\downarrow$)} & {49.58 (4.33$\uparrow$)} & {79.43 (2.09$\downarrow$)} & {48.10 (3.99$\uparrow$)} \\
    \cmidrule{1-9}
    {GrADS} & {74.49 (2.37$\downarrow$)} & {65.01 (3.37$\uparrow$)} & {80.22 (1.30$\downarrow$)} & {64.08 (1.24$\uparrow$)} & {75.68 (1.18$\downarrow$)} & {51.77 (6.52$\uparrow$)} & {80.79 (0.73$\downarrow$)} & {49.41 (5.30$\uparrow$)} \\
    {OGS} & \underline{75.79 (1.07$\downarrow$)} & {63.73 (2.09$\uparrow$)} & \underline{80.69 (0.83$\downarrow$)} & {62.58 (-0.26$\uparrow$)} & {75.24 (1.62$\downarrow$)} & {51.42 (6.17$\uparrow$)} & {79.49 (2.03$\downarrow$)} & {48.05 (3.94$\uparrow$)} \\
    \cmidrule{1-9}
    {Selective DPO} & {71.68 (5.18$\downarrow$)} & \textbf{67.28 (5.64$\uparrow$)} & {79.33 (2.19$\downarrow$)} & {66.21 (3.37$\uparrow$)} & {71.45 (5.41$\downarrow$)} & \underline{55.78 (10.53$\uparrow$)} & {78.10 (3.42$\downarrow$)} & \textbf{53.02 (8.91$\uparrow$)} \\
    {BeeS} & {73.32 (3.54$\downarrow$)} & {66.00 (4.36$\uparrow$)} & {80.28 (1.24$\downarrow$)} & {65.70 (2.86$\uparrow$)} & {73.86 (3.00$\downarrow$)} & {54.62 (9.37$\uparrow$)} & {80.39 (1.13$\downarrow$)} & {50.63 (6.52$\uparrow$)} \\
    {PD} & {72.67 (4.19$\downarrow$)} & \underline{66.25 (4.61$\uparrow$)} & {80.07 (1.45$\downarrow$)} & \textbf{66.30  (3.46$\uparrow$)} & {73.60 (3.26$\downarrow$)} & {54.97 (9.72$\uparrow$)} & {80.80 (0.72$\downarrow$)} & {49.54 (5.43$\uparrow$)} \\
    \cmidrule{1-9}
    {BALIGN (Ours)} & \textbf{76.00 (0.86$\downarrow$)} & \underline{66.25 (4.61$\uparrow$)} & \textbf{81.14 (0.38$\downarrow$)} & \underline{66.23 (3.39$\uparrow$)} & \textbf{75.93 (0.93$\downarrow$)} & \textbf{56.07 (10.82$\uparrow$)} & \textbf{81.39 (0.13$\downarrow$)} & \underline{52.48 (8.37$\uparrow$)} \\
    \bottomrule
    \end{tabular}
    \vspace{-0.25cm}
    \caption{General capability and alignment results on the HH-RLHF dataset, evaluated separately for helpfulness and harmlessness. We use Llama-3.1-8B-Instruct and Qwen2.5-7B-Instruct models as base reference models. The best and second-best results are highlighted in bold and underline, respectively, excluding the base model performance.}
  \label{tbl:main_exp}
  \vspace{-0.3cm}
\end{table*}

\begin{figure*}[t]
    \centering
    \begin{subfigure}[t]{\textwidth}
        \centering
        \includegraphics[width=0.5\linewidth]{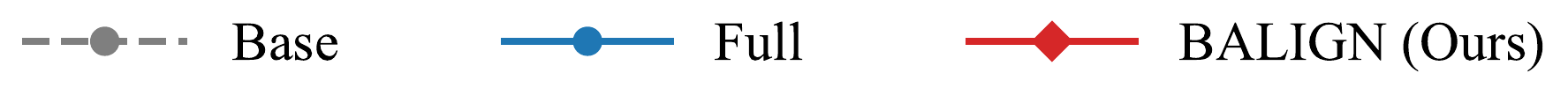} 
        \label{fig:radar_legend}
    \end{subfigure}
    \begin{subfigure}[t]{0.246\textwidth}
        \includegraphics[width=\linewidth]{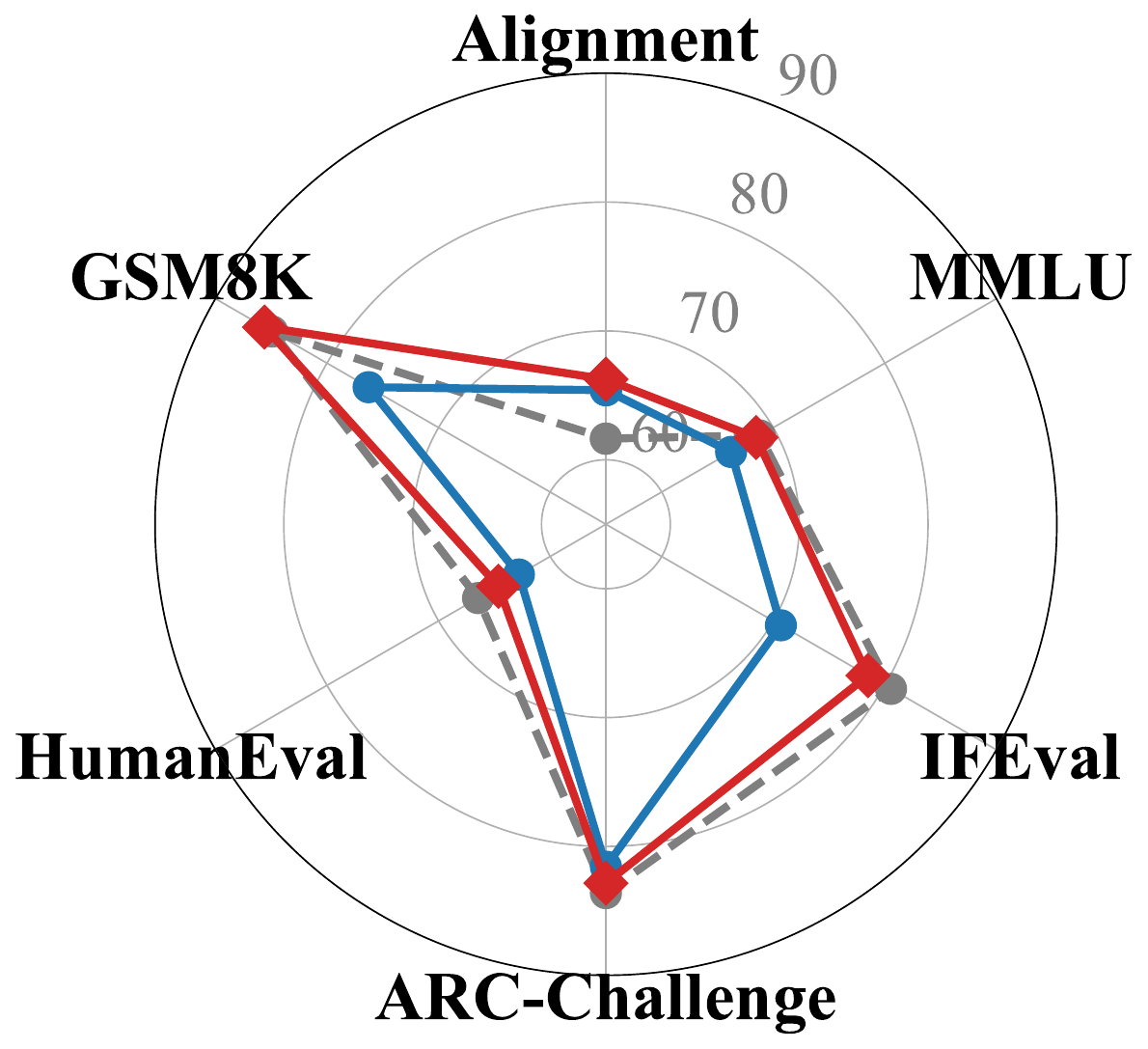}
        \caption{Helpfulness -- Llama.}
        \label{fig:radar1}
    \end{subfigure}
    \begin{subfigure}[t]{0.246\textwidth}
        \includegraphics[width=\linewidth]{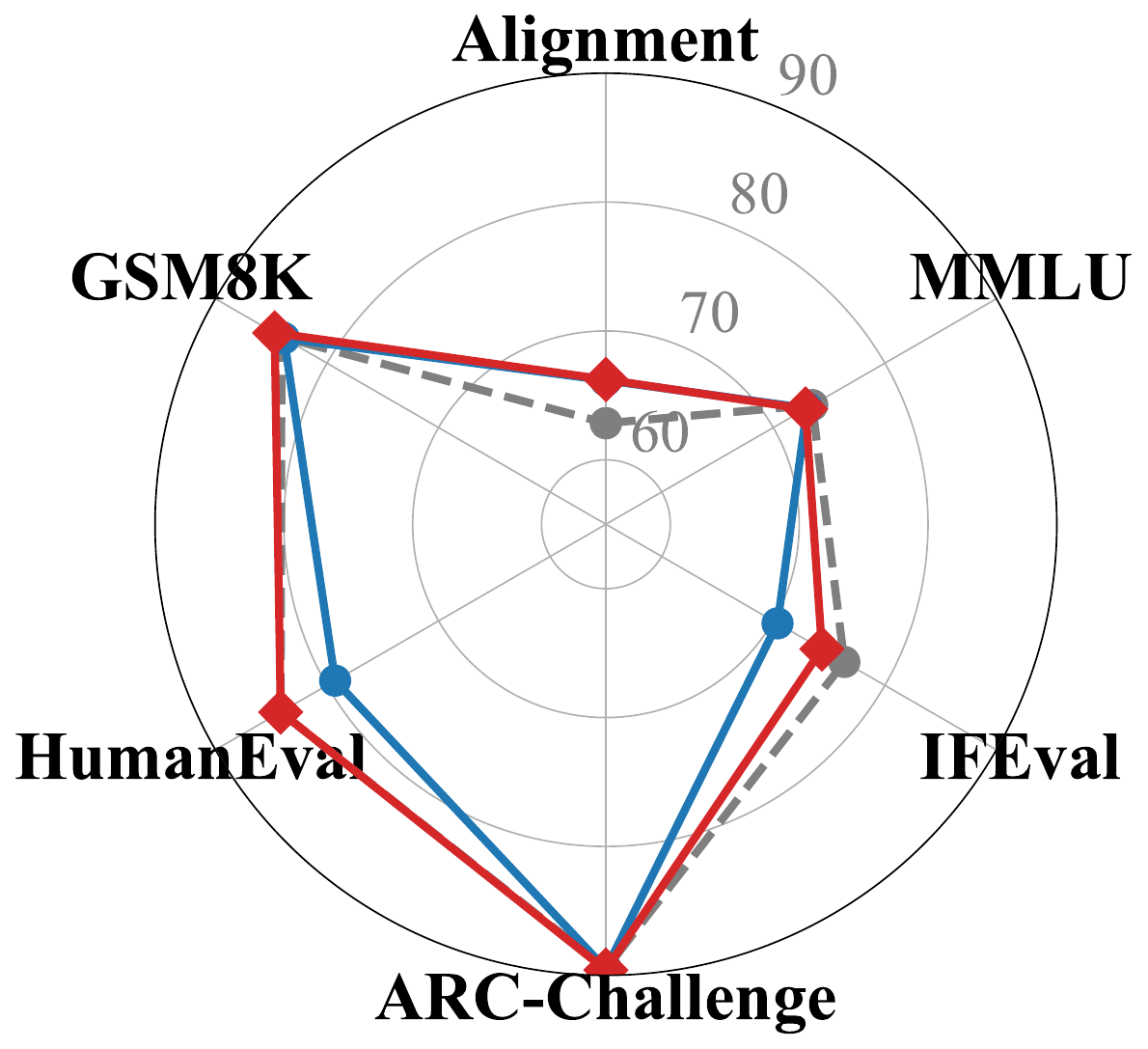}
        \caption{Helpfulness -- Qwen.}
        \label{fig:radar2}
    \end{subfigure}
    \begin{subfigure}[t]{0.246\textwidth}
        \includegraphics[width=\linewidth]{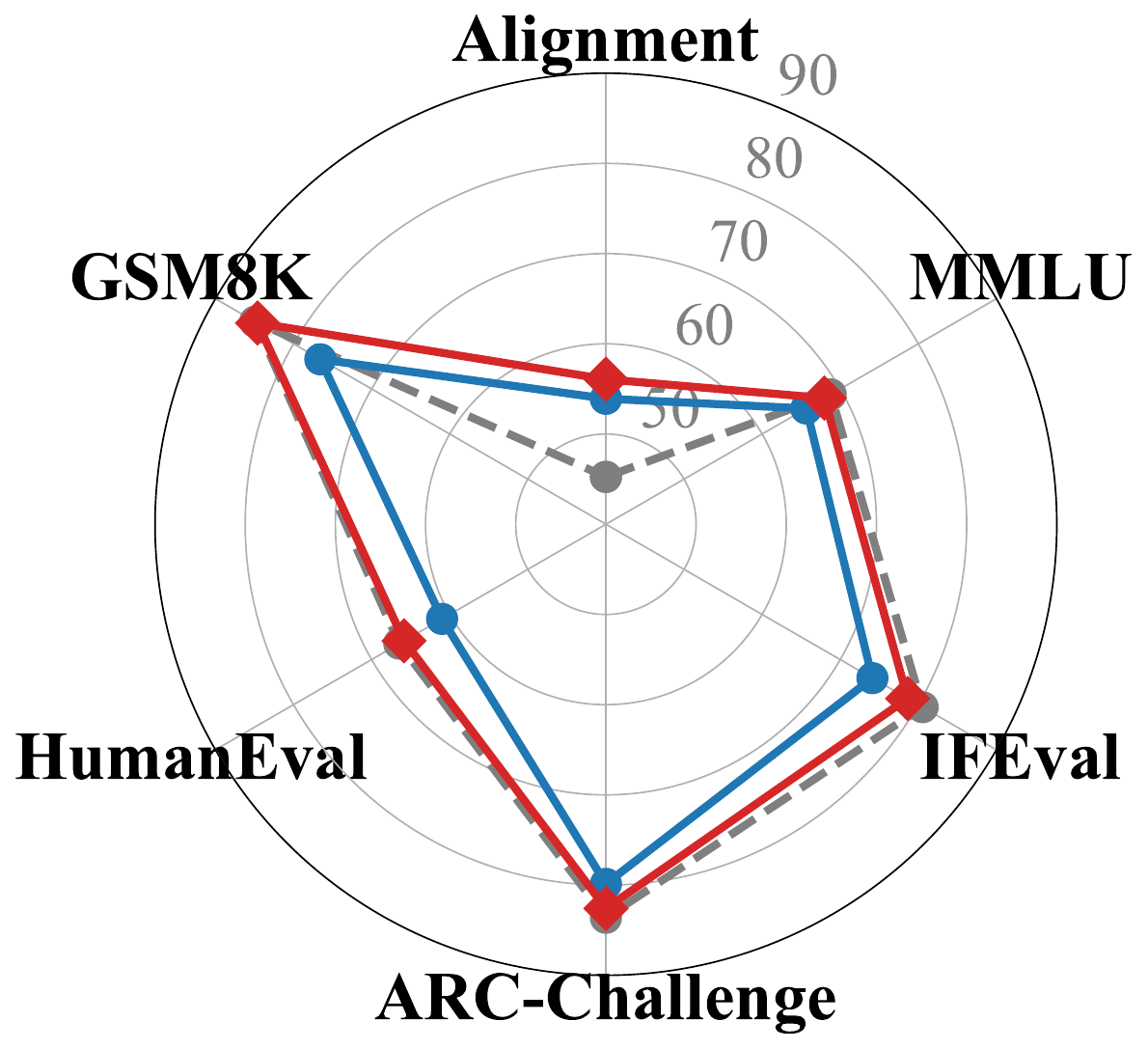}
        \caption{Harmlessness -- Llama.}
        \label{fig:radar3}
    \end{subfigure}
    \begin{subfigure}[t]{0.246\textwidth}
        \includegraphics[width=\linewidth]{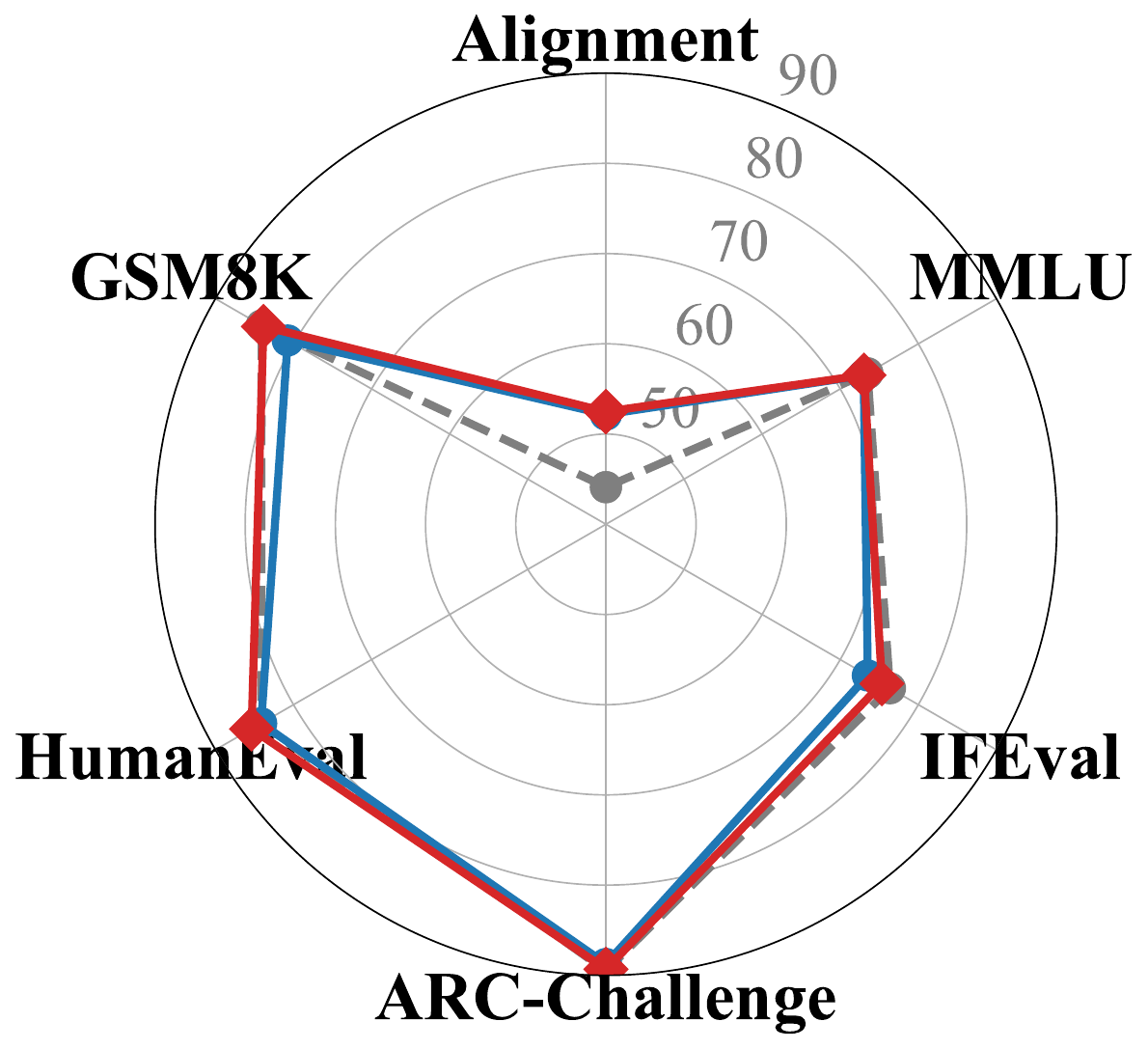}
        \caption{Harmlessness -- Qwen.}
        \label{fig:radar4}
    \end{subfigure}
    \vspace{-0.2cm}
    \caption{Detailed general capabilities and alignment of LLMs. \method{} successfully preserves the base model's diverse general capabilities while achieving alignment performance on par with training on the full preference dataset.}
    \label{fig:radar}
    \vspace{-0.45cm}
\end{figure*}

\section{Experiments}

We conduct experiments to evaluate \method{} and provide a detailed analysis of the results. 
We provide more detailed experimental settings and results in Appendix.

\paragraph{Metrics.}

We evaluate the methods along two axes: general capability and alignment, which correspond to stability and plasticity, respectively. We first measure the average absolute performance of the models on general capability ($General$) and alignment ($Align$) using Eqs.~(\ref{eq:general}) and~(\ref{eq:alignment}), respectively. We then quantify the performance shifts relative to the base reference model, defining these differences as catastrophic forgetting ($F$) and alignment gain ($R$) using Eqs.~(\ref{eq:forgetting}) and~(\ref{eq:alignment_gain}). 

\paragraph{Datasets.}

For alignment, we use the HH-RLHF\,\cite{DBLP:journals/corr/abs-2204-05862} dataset to train the model for helpfulness and harmlessness. We utilize separate training and test splits of the dataset for training and evaluation. 
For general capability, we use a total of five datasets following the benchmark evaluation settings\,\cite{DBLP:journals/corr/abs-2310-06825, DBLP:journals/corr/abs-2407-21783, DBLP:journals/corr/abs-2412-15115}, including MMLU\,\cite{DBLP:conf/iclr/HendrycksBBZMSS21}, IFEval\,\cite{DBLP:journals/corr/abs-2311-07911}, ARC-Challenge\,\cite{DBLP:journals/corr/abs-1803-05457}, HumanEval\,\cite{DBLP:journals/corr/abs-2107-03374}, and GSM8K\,\cite{DBLP:journals/corr/abs-2110-14168}. These datasets cover general knowledge, instruction following, reasoning, code, and math, which constitute the fundamental capabilities of LLMs. We reserve a subset of the general capability datasets as a validation set for required baselines, while the remainder serves as a test set.

\paragraph{Models.}

We adopt Llama-3.1-8B-Instruct\,\cite{DBLP:journals/corr/abs-2407-21783} and Qwen2.5-7B-Instruct\,\cite{DBLP:journals/corr/abs-2412-15115} as the base reference models for our experiments. We select instruction-tuned models in the 7B--8B parameter range as they offer an optimal balance between robust base capabilities and tractability within our GPU computational budget. 

\paragraph{Baselines.}

We compare \method{} with four categories of data selection baselines. 
Within this taxonomy, \method{} is a stability- and plasticity-aware alignment method.

\begin{itemize}[leftmargin=*]
\setlength\itemsep{0cm}
\item {\bf Naive methods}: {\it Full} and {\it Random}.
\item {\bf Plasticity-aware SFT methods}: {\it DSIR}\,\cite{DBLP:conf/nips/XieS0L23}, {\it LESS}\,\cite{DBLP:conf/icml/XiaMGA024}, {\it TSDS}\,\cite{DBLP:conf/nips/LiuKR24}, and {\it NICE}\,\cite{DBLP:conf/icml/0001L0KFL25}.
\item {\bf Stability- and Plasticity-aware SFT methods}: {\it GrADS}\,\cite{DBLP:journals/corr/abs-2511-08620} and {\it OGS}\,\cite{DBLP:journals/corr/abs-2602-06359}.
\item {\bf Plasticity-aware Alignment methods}: {\it Selective DPO}\,\cite{DBLP:conf/icml/GaoL0XZ025}, {\it BeeS}\,\cite{DBLP:conf/nips/DengZAFWH25}, and {\it PD}\,\cite{zhang2026dataselectionllmalignment}.
\end{itemize}

\paragraph{Hyperparameters.}

We perform a grid search on the held-out validation set to find the optimal hyperparameters. For the reference margin-based risk score, we set $\alpha$ to 0.25 and 0.15 for the Llama-3.1-8B-Instruct and Qwen2.5-7B-Instruct models, respectively. For the length difference-based risk score, we set $\gamma$ to 0.5 and 0.25, respectively. When formulating the composite risk score, we set $\lambda$ to 1 and 2, respectively. 

\begin{figure*}[t]
    \begin{subfigure}[t]{0.2465\textwidth}
        \includegraphics[width=\linewidth]{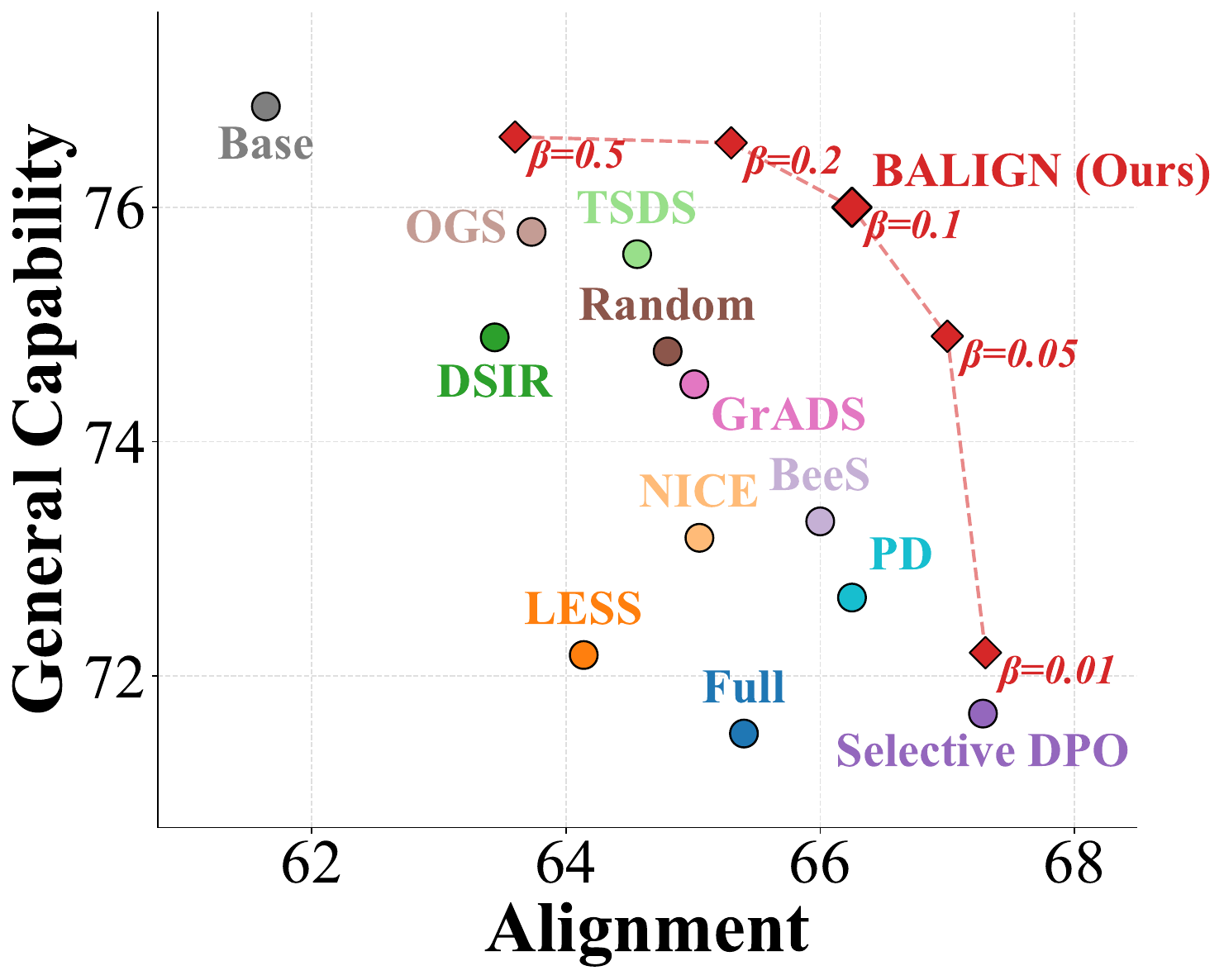}
        \caption{Helpfulness -- Llama.}
        \label{fig:pareto1}
    \end{subfigure}
    \begin{subfigure}[t]{0.2465\textwidth}
        \includegraphics[width=\linewidth]{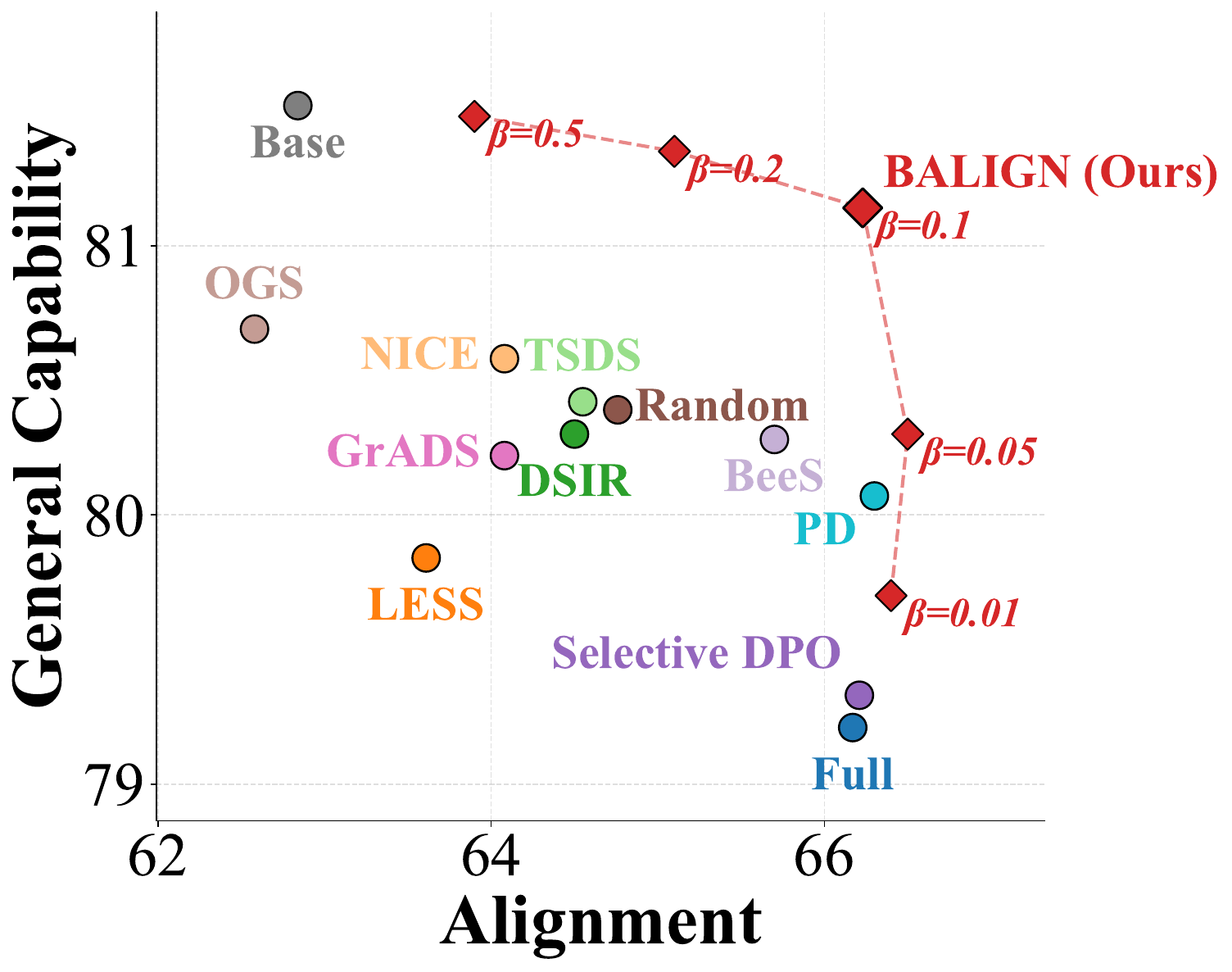}
        \caption{Helpfulness -- Qwen.}
        \label{fig:pareto2}
    \end{subfigure}
    \begin{subfigure}[t]{0.2465\textwidth}
        \includegraphics[width=\linewidth]{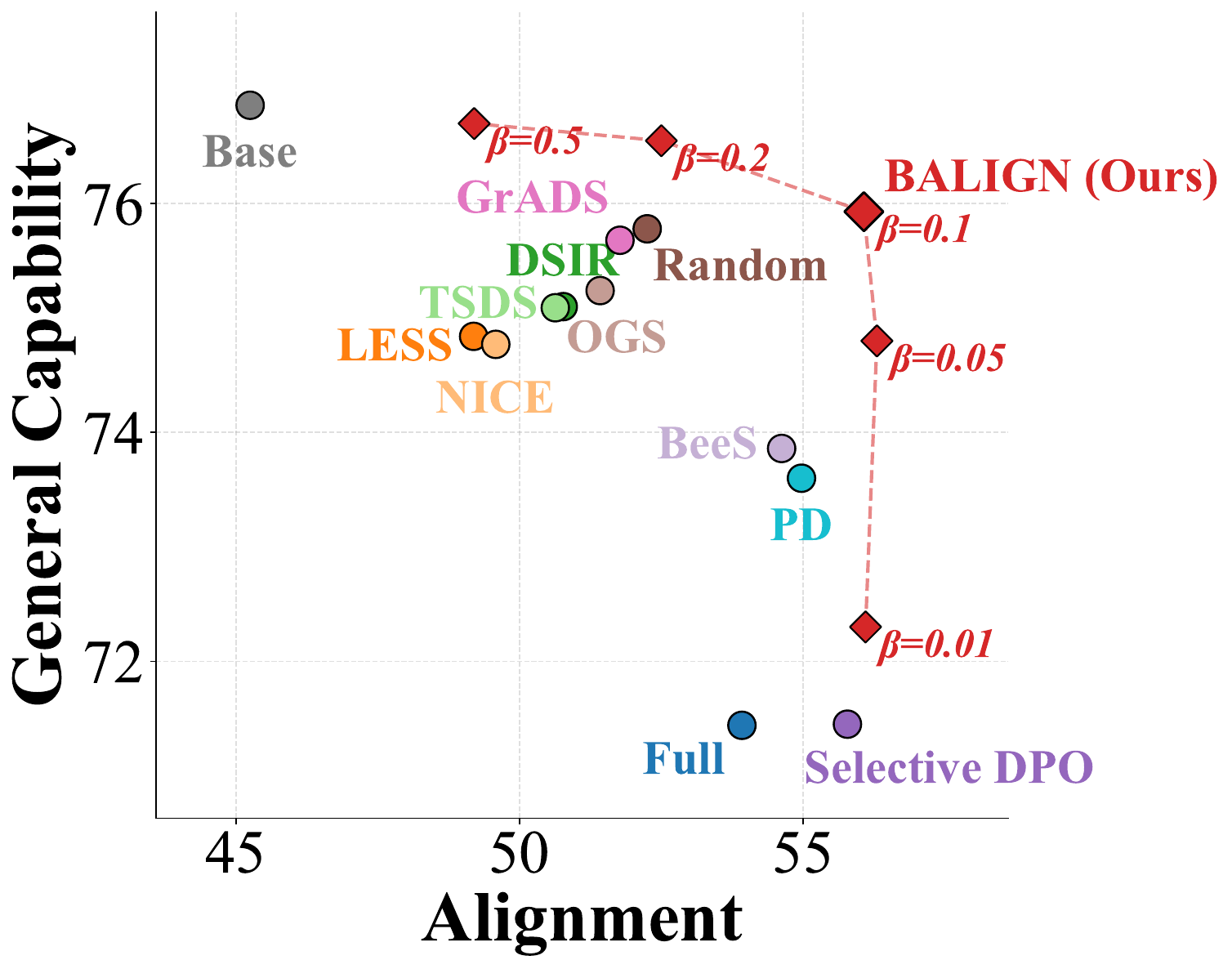}
        \caption{Harmlessness -- Llama.}
        \label{fig:pareto3}
    \end{subfigure}
    \begin{subfigure}[t]{0.2465\textwidth}
        \includegraphics[width=\linewidth]{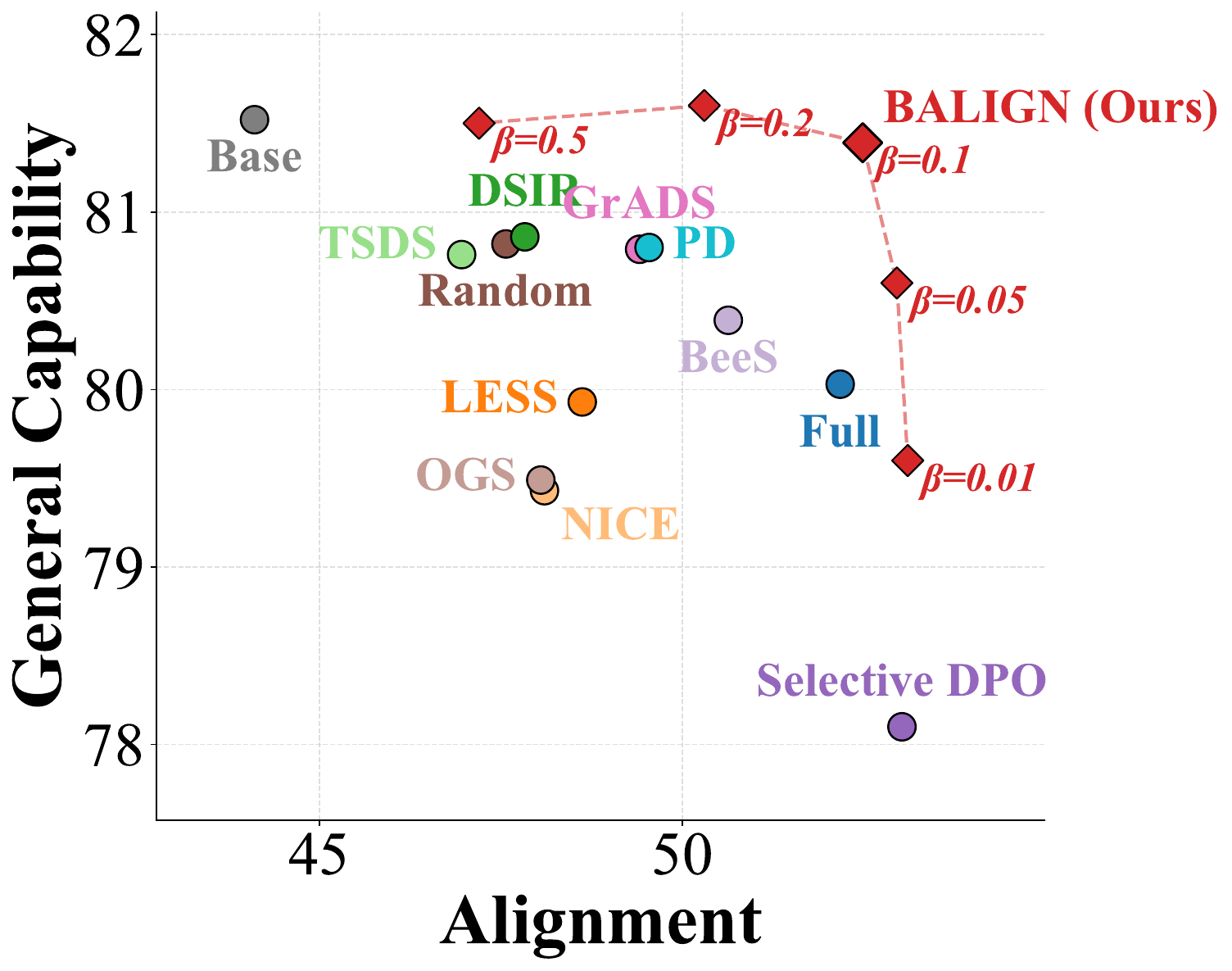}
        \caption{Harmlessness -- Qwen.}
        \label{fig:pareto4}
    \end{subfigure}
    \vspace{-0.2cm}
    \caption{Pareto frontier of \method{} against baselines on the HH-RLHF dataset.}
    \label{fig:main_control}
    \vspace{-0.25cm}
\end{figure*}

\begin{table*}[t]
    \small
    \setlength{\tabcolsep}{2.8pt}
    \centering
    \begin{tabular}{l|cc|cc|cc|cc}
        \toprule
        {Method} & \multicolumn{4}{c|}{\sf Helpfulness} & \multicolumn{4}{c}{\sf Harmlessness} \\
        \cmidrule{1-9}
        {} & \multicolumn{2}{c|}{\sf Llama-3.1-8B-Instruct} & \multicolumn{2}{c|}{\sf Qwen2.5-7B-Instruct} & \multicolumn{2}{c|}{\sf Llama-3.1-8B-Instruct} & \multicolumn{2}{c}{\sf Qwen2.5-7B-Instruct} \\
        \cmidrule{1-9}
        {} & {General (F$\downarrow$)} & {Align (R$\uparrow$)} & {General (F$\downarrow$)} & {Align (R$\uparrow$)} & {General (F$\downarrow$)} & {Align (R$\uparrow$)} & {General (F$\downarrow$)} & {Align (R$\uparrow$)} \\
        \midrule
        {Base} & {76.86 (--)} & {61.64 (--)} & {81.52 (--)} & {62.84 (--)} & {76.86 (--)} & {45.25 (--)} & {81.52 (--)} & {44.11 (--)} \\
        \cmidrule{1-9}
        {W/o $s_{\Delta p_{\mathrm{ref}}}$} & {75.18 (1.68$\downarrow$)} & {65.01 (3.37$\uparrow$)} & {80.50 (1.02$\downarrow$)} & {65.34 (2.50$\uparrow$)} & {74.67 (2.19$\downarrow$)} & {53.04 (7.79$\uparrow$)} & {79.94 (1.58$\downarrow$)} & {48.70 (4.59$\uparrow$)} \\
        {W/o $s_{\Delta\ell}$} & {75.59 (1.27$\downarrow$)} & {65.95 (4.31$\uparrow$)} & {79.98 (1.54$\downarrow$)} & {65.56 (2.72$\uparrow$)} & {74.27 (2.59$\downarrow$)} & {53.34 (8.09$\uparrow$)} & {80.19 (1.33$\downarrow$)} & {51.07 (6.96$\uparrow$)} \\
        {W/o $s_{\tau}$} & {75.78 (1.08$\downarrow$)} & {64.03 (2.39$\uparrow$)} & {80.44 (1.08$\downarrow$)} & {64.89 (2.05$\uparrow$)} & {75.30 (1.56$\downarrow$)} & {51.11 (5.86$\uparrow$)} & {80.80 (0.72$\downarrow$)} & {47.96 (3.85$\uparrow$)} \\
        \cmidrule{1-9}
        {BALIGN (Ours)} & \textbf{76.00 (0.86$\downarrow$)} & \textbf{66.25 (4.61$\uparrow$)} & \textbf{81.14 (0.38$\downarrow$)} & \textbf{66.23 (3.39$\uparrow$)} & \textbf{75.93 (0.93$\downarrow$)} & \textbf{56.07 (10.82$\uparrow$)} & \textbf{81.39 (0.13$\downarrow$)} & \textbf{52.48 (8.37$\uparrow$)} \\
        \bottomrule
    \end{tabular}
    \vspace{-0.2cm}
    \caption{Ablation study of \method{} on the HH-RLHF dataset, evaluated separately for helpfulness and harmlessness.
    }
    \label{tbl:ablation_exp}
    \vspace{-0.4cm}
\end{table*}

\subsection{General Capability and Alignment Results}
\label{subsec:results}
\vspace{-0.01cm}
We compare \method{} against four different categories of data selection baselines as shown in Table~\ref{tbl:main_exp}. We also present detailed capability results for each task in Figure~\ref{fig:radar} and Appendix. To ensure a fair comparison, all methods perform data selection subject to a fixed data budget, defined by a given data sampling ratio. Following standard practice in prior LLM data selection studies\,\cite{DBLP:conf/icml/XiaMGA024, DBLP:conf/nips/LiuKR24, DBLP:conf/icml/0001L0KFL25}, we set the default data sampling ratio to 5\% of the full dataset. 

Overall, \method{} consistently achieves the optimal balance between general capability and alignment by minimizing catastrophic forgetting while preserving alignment signals. Specifically, \method{} achieves the best performance in general capability and the best or second-best performance in alignment, with only a marginal gap compared to the top performer. Since plasticity-aware alignment methods focus solely on alignment without considering catastrophic forgetting, outperforming them on the alignment axis is inherently difficult. In contrast, as the other data selection baselines are designed primarily for SFT data, they exhibit limitations when applied to alignment setups with preference data. 
These results demonstrate that \method{} serves as a highly effective stability- and plasticity-aware alignment method.

\subsection{Controllability of Stability and Plasticity}

We next investigate the controllability of \method{} by leveraging the hyperparameter $\beta$ in the DPO objective, which governs the strength of the implicit Kullback-Leibler (KL) divergence penalty relative to the base reference model. 
By systematically varying $\beta \in \{0.01, 0.05, 0.1, 0.2, 0.5\}$, we analyze how effectively \method{} allows practitioners to modulate the balance between stability and plasticity as shown in Figure~\ref{fig:main_control}. Overall, reducing $\beta$ diminishes the regularization effect, thereby generally prioritizing plasticity, whereas increasing $\beta$ enforces stronger regularization to preserve stability. Across this spectrum, \method{} consistently dominates the baselines and traces the Pareto frontier, distinctly occupying the optimal top-right region of the trade-off space, with the default $\beta{=}0.1$ providing a strong balance.

\subsection{Ablation Study}

To verify the effectiveness of each risk score in \method{}, we perform an ablation study as shown in Table~\ref{tbl:ablation_exp}. We evaluate the impact of individual components by removing each of the three risk scores from the composite risk score. The results demonstrate that the exclusion of any single risk score consistently compromises the equilibrium between model stability and plasticity. For example, omitting either $s_{\Delta p_{\mathrm{ref}}}$ or $s_{\Delta\ell}$ exacerbates catastrophic forgetting, leading to a pronounced deterioration in general capabilities. Conversely, excluding $s_{\tau}$ substantially impedes the model's capacity to maximize alignment gains. These findings substantiate that the three risk scores operate in a complementary manner, exhibiting a strong synergistic effect upon integration.


\subsection{Computation Time Analysis}

\begin{figure}[t]
    \centering
    \includegraphics[width=0.95\linewidth]{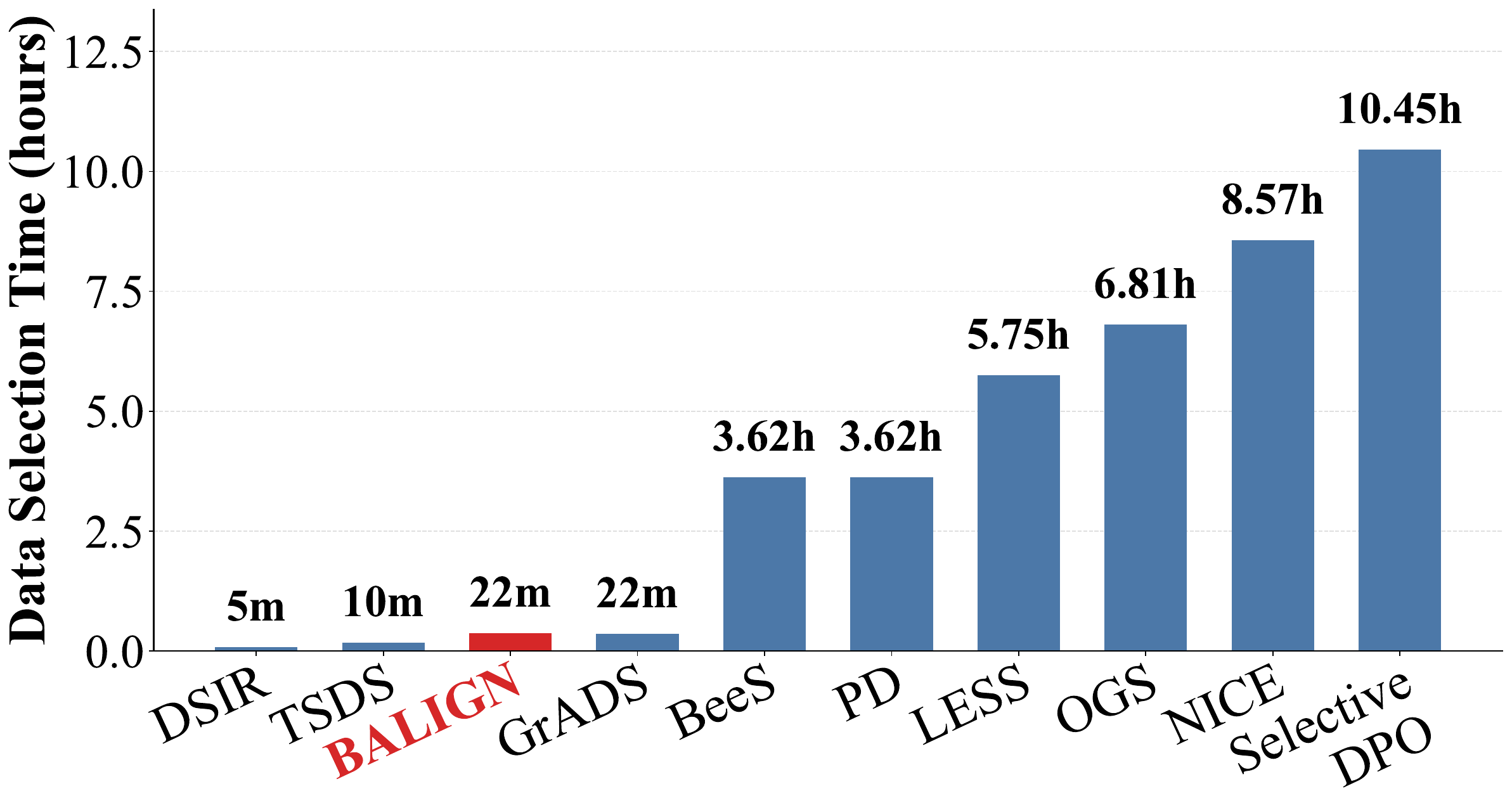}
    \vspace{-0.1cm}
    \caption{Computation time of data selection baselines.}
    \label{fig:time}
    \vspace{-0.4cm}
\end{figure}

To demonstrate the efficiency of our approach, we conduct a comparative analysis of the computational overhead incurred during the data selection phase. Specifically, we measure and compare the total computation time required by \method{} and the baseline methods to select the optimal subset from the full preference dataset as shown in Figure~\ref{fig:time}. Gradient-based methods (e.g., LESS, NICE, OGS) and alignment methods (e.g., Selective DPO, BeeS, PD) incur prohibitive computational overhead due to expensive per-sample gradient computations or auxiliary model training. Meanwhile, GrADS achieves computational efficiency by compressing per-sample gradients into scalar norms and operating without validation targets. While statistical baselines (e.g., DSIR, TSDS) are highly efficient, they inherently ignore model-specific preference signals, yielding suboptimal alignment gains. In contrast, \method{} effectively bridges this gap by computing three lightweight risk scores that require at most a single forward pass over the base reference model.

\section{Conclusion}

In this work, we address the critical stability-plasticity dilemma in LLM alignment by introducing \method{}, a highly efficient, data-centric framework designed to jointly optimize model stability and plasticity. By analyzing the preference optimization gradient, we demonstrate that catastrophic forgetting is exacerbated by preference samples exhibiting extreme reference margins and significant length asymmetries, whereas lexical overlap with general capability domains governs the model's capacity to maximize alignment gains. \method{} seamlessly integrates these insights into a unified composite risk score, enabling the effective removal of high-risk samples prior to optimization. Our comprehensive evaluations confirm that this balanced data selection strategy not only establishes a superior trade-off between capability retention and preference alignment, but also incurs a significantly lower computational overhead than existing baselines. Therefore, \method{} provides a scalable and practical foundation for preprocessing preference data, paving the way for developing aligned LLMs without compromising their foundational versatility.



\bibliography{aaai2027}


\appendix
\onecolumn

\section{Appendix}

\subsection{More Details on Empirical Analysis}

We provide more details on empirical analysis in Table~\ref{tab:bucket_results}. The results demonstrate that our proposed risk scores capture complementary dimensions of model degradation. Specifically, $s_{\Delta p_{\mathrm{ref}}}$ and $s_{\Delta\ell}$ serve as primary indicators of catastrophic forgetting. As the risk increases from $\mathcal{B}^{1}$ to $\mathcal{B}^{5}$, the forgetting metric exhibits a pronounced upward trajectory for $\Delta p_{\mathrm{ref}}$ and $\Delta\ell$, reaching 3.48\% and 2.19\%, respectively. In contrast, $s_{\tau}$ primarily captures the failure of preference learning. For this feature, the highest-risk bucket yields a significantly diminished alignment gain while maintaining a relatively stable forgetting rate. These findings empirically substantiate our theoretical formulation that $s_{\Delta p_{\mathrm{ref}}}$ and $s_{\Delta\ell}$ effectively identify samples prone to destroying pre-trained knowledge, while $s_{\tau}$ isolates samples detrimental to the acquisition of human preferences. This distinct yet complementary behavior underscores the necessity of integrating all three features for a robust alignment framework.



\subsection{More Details on Experimental Settings}

We provide more details on experimental settings. All training and evaluation are run on Intel Xeon Silver 4210R CPUs and NVIDIA RTX A6000 GPUs. 

\paragraph{Metrics.}

Since both general capability and alignment are crucial, we evaluate the overall effectiveness of each method based on its optimal balance between the two axes. 
We note that maximizing alignment gain ($R$) while minimizing catastrophic forgetting ($F$) represents desirable model performance. During evaluation, we employ greedy decoding for all generative benchmarks and log-likelihood scoring for multiple-choice tasks, thereby eliminating stochasticity from the reported metrics.

\paragraph{Datasets.}

For alignment, we use the HH-RLHF\,\cite{DBLP:journals/corr/abs-2204-05862} dataset to train the model for helpfulness and harmlessness. We utilize separate training and test splits of the dataset for training and evaluation. 
Specifically, the helpfulness and harmlessness subsets contain 43,835 and 42,537 training samples with 2,354 and 2,312 test samples, respectively.

\paragraph{Models.}

We adopt Llama-3.1-8B-Instruct\,\cite{DBLP:journals/corr/abs-2407-21783} and Qwen2.5-7B-Instruct\,\cite{DBLP:journals/corr/abs-2412-15115} as the base reference models for our experiments. We select instruction-tuned models in the 7B--8B parameter range as they offer an optimal balance between robust base capabilities and tractability within our GPU computational budget. 
Since our methodology requires parameter updates during training, we cannot use closed-weight proprietary models.

\paragraph{Baselines.}

We compare \method{} with four categories of data selection baselines. 
These categories are chosen to provide a comprehensive comparison, covering basic selection methods, objective-driven methods for supervised fine-tuning (SFT), and recent methods specifically tailored for preference alignment. 
To adapt the SFT-specific baselines to preference datasets, we implement their data selection process by treating the chosen response in each preference pair as the ground-truth target. 
Within this taxonomy, \method{} is a stability- and plasticity-aware alignment method.

\begin{itemize}[leftmargin=*]
\setlength\itemsep{0cm}
\item {\bf Naive methods}: {\it Full} uses the entire preference dataset without selection; and {\it Random} draws a uniformly random subset of the preference dataset.
\item {\bf Plasticity-aware SFT methods}: {\it DSIR}\,\cite{DBLP:conf/nips/XieS0L23} performs importance resampling with hashed $n$-gram features to match the target-task distribution; {\it LESS}\,\cite{DBLP:conf/icml/XiaMGA024} selects samples whose low-rank gradient projections align with the target-task gradient direction; {\it TSDS}\,\cite{DBLP:conf/nips/LiuKR24} selects samples whose dense embeddings collectively match the target-task distribution while penalizing redundancy; and {\it NICE}\,\cite{DBLP:conf/icml/0001L0KFL25} extends LESS by casting selection as a policy-gradient optimization problem.
\item {\bf Stability- and Plasticity-aware SFT methods}: {\it GrADS}\,\cite{DBLP:journals/corr/abs-2511-08620} applies kernel density estimation on embedding- and logit-gradient magnitudes to jointly favor in-distribution samples for both target and anchor; and {\it OGS}\,\cite{DBLP:journals/corr/abs-2602-06359} retains samples whose projected gradients are orthogonal to a capability-anchor gradient.
\item {\bf Plasticity-aware Alignment methods}: {\it Selective DPO}\,\cite{DBLP:conf/icml/GaoL0XZ025} trains auxiliary DPO references on data partitions and retains samples with the lowest validation loss; {\it BeeS}\,\cite{DBLP:conf/nips/DengZAFWH25} applies Bayesian aggregation to implicit and external reward margins for consensus-based selection; and {\it PD}\,\cite{zhang2026dataselectionllmalignment} trains proxy reward models on multiple fine-grained sub-preferences and selects samples with the strongest cross-aspect consensus.
\end{itemize}


\paragraph{Hyperparameters.}

For DPO training, we set the effective batch size to 128, use a learning rate of $5\times 10^{-6}$ with a cosine schedule (10\% warmup) and a sigmoid preference loss with $\beta{=}0.1$, and train for 3 epochs in \texttt{bfloat16} precision under DeepSpeed ZeRO-3\,\cite{DBLP:conf/sc/RajbhandariRRH20} with full-parameter fine-tuning.

\clearpage
\newpage


\subsection{More Details on General Capability and Alignment Results}

We provide more details on general capability and alignment results as shown in Tables~\ref{tab:hh-rlhf-baselines}--\ref{tab:hh-rlhf-harmless-qwen}.

\begin{table*}[h]
\small
\centering
\setlength{\tabcolsep}{4.65pt}
\centering
\begin{tabular}{l|ccccc|cc|cc}
\toprule
Method & MMLU & IFEval & ARC-Challenge & HumanEval & GSM8K & General & Forgetting & Alignment & Gain \\
\midrule
Base (Llama-3.1-8B-Instruct) & 68.78 & 80.55 & 83.62 & 66.46 & 84.91 & 76.86 & -- & 61.64 & -- \\
\midrule
Full & 66.22 & 70.68 & 81.57 & 62.80 & 76.27 & 71.51 & 5.35 & 65.40 & 3.76 \\
Random & 68.22 & 80.20 & 83.11 & 59.15 & 83.17 & 74.77 & 2.09 & 64.80 & 3.16 \\ 
\midrule
DSIR & 68.24 & 78.60 & 82.94 & 62.80 & 81.88 & 74.89 & 1.97 & 63.44 & 1.80 \\
LESS & 68.05 & 72.99 & 82.51 & 60.98 & 76.35 & 72.18 & 4.68 & 64.14 & 2.50 \\
TSDS & 68.78 & 79.67 & 82.34 & 64.02 & 83.17 & 75.60 & 1.26 & 64.56 & 2.92 \\
NICE & 67.99 & 78.33 & 81.83 & 57.93 & 79.83 & 73.18 & 3.68 & 65.05 & 3.41 \\
\midrule
GrADS & 68.17 & 77.47 & 82.51 & 64.02 & 80.29 & 74.49 & 2.37 & 65.01 & 3.37 \\
OGS & 68.57 & 79.55 & 83.28 & 63.41 & 84.15 & \underline{75.79} & \underline{1.07} & 63.73 & 2.09 \\
\midrule
Selective DPO & 67.98 & 68.10 & 83.28 & 59.76 & 79.30 & 71.68 & 5.18 & \textbf{67.28} & \textbf{5.64} \\
BeeS & 68.32 & 73.31 & 83.62 & 60.37 & 80.97 & 73.32 & 3.54 & 66.00 & 4.36 \\
PD & 68.33 & 74.35 & 83.53 & 57.32 & 79.83 & 72.67 & 4.19 & \underline{66.25} & \underline{4.61} \\ 
\midrule
BALIGN (Ours) & 68.46 & 78.48 & 82.85 & 64.63 & 85.60 & \textbf{76.00} & \textbf{0.86} & \underline{66.25} & \underline{4.61} \\
\bottomrule
\end{tabular}
\vspace{-0.2cm}
\caption{Detailed general capability and alignment results for helpfulness on the HH-RLHF dataset, using Llama-3.1-8B-Instruct as the base reference model. The best and second-best results are highlighted in bold and underline, respectively, excluding the base model performance.}
\label{tab:hh-rlhf-baselines}
\end{table*}

\begin{table*}[h]
\small
\centering
\setlength{\tabcolsep}{4.85pt}
\centering 
\begin{tabular}{l|ccccc|cc|cc}
\toprule 
Method & MMLU & IFEval & ARC-Challenge & HumanEval & GSM8K & General & Forgetting & Alignment & Gain \\ 
\midrule
Base (Qwen2.5-7B-Instruct) & 73.49 & 76.37 & 89.51 & 84.15 & 84.08 & 81.52 & -- & 62.84 & -- \\ 
\midrule 
Full & 73.02 & 70.38 & 89.51 & 79.27 & 83.85 & 79.21 & 2.31 & 66.17 & 3.33 \\ 
Random & 73.37 & 74.33 & 89.16 & 81.10 & 84.00 & 80.39 & 1.13 & 64.76 & 1.92 \\ 
\midrule 
DSIR & 72.85 & 74.00 & 89.51 & 78.66 & 86.50 & 80.30 & 1.22 & 64.50 & 1.66 \\ 
LESS & 73.47 & 72.37 & 89.33 & 78.66 & 85.37 & 79.84 & 1.68 & 63.61 & 0.77 \\ 
TSDS & 73.34 & 72.55 & 89.25 & 83.54 & 83.40 & 80.42 & 1.10 & 64.55 & 1.71 \\ 
NICE & 73.86 & 72.85 & 89.42 & 79.27 & 87.49 & 80.58 & 0.94 & 64.08 & 1.24 \\ 
\midrule 
GrADS & 73.34 & 74.77 & 89.51 & 79.88 & 83.62 & 80.22 & 1.30 & 64.08 & 1.24 \\ 
OGS & 72.97 & 75.84 & 89.68 & 82.93 & 82.03 & \underline{80.69} & \underline{0.83} & 62.58 & -0.26 \\ 
\midrule 
Selective DPO & 73.47 & 72.35 & 89.59 & 77.44 & 83.78 & 79.33 & 2.19 & 66.21 & 3.37 \\ 
BeeS & 73.24 & 72.76 & 89.85 & 80.49 & 85.06 & 80.28 & 1.24 & 65.70 & 2.86 \\ 
PD & 73.37 & 72.95 & 89.76 & 78.66 & 85.60 & 80.07 & 1.45 & \textbf{66.30} & \textbf{3.46} \\ 
\midrule 
BALIGN (Ours) & 72.89 & 74.36 & 89.59 & 84.15 & 84.69 & \textbf{81.14} & \textbf{0.38} & \underline{66.23} & \underline{3.39} \\ 
\bottomrule 
\end{tabular}
\vspace{-0.2cm}
\caption{Detailed general capability and alignment results for helpfulness on the HH-RLHF dataset, using Qwen2.5-7B-Instruct as the base reference model. The best and second-best results are highlighted in bold and underline, respectively, excluding the base model performance.}
\label{tab:hh-rlhf-baselines-qwen}
\end{table*}

\clearpage
\newpage

\begin{table*}[h]
  \small
  \centering
  \setlength{\tabcolsep}{4.52pt}
  \centering
  \begin{tabular}{l|ccccc|cc|cc}
  \toprule
  Method & MMLU & IFEval & ARC-Challenge & HumanEval & GSM8K & General & Forgetting & Alignment & Gain \\
  \midrule
  Base (Llama-3.1-8B-Instruct) & 68.78 & 80.55 & 83.62 & 66.46 & 84.91 & 76.86 & -- & 45.25 & -- \\
  \midrule
  Full & 65.66 & 74.12 & 79.86 & 60.98 & 76.57 & 71.44 & 5.42 & 53.92 & 8.67 \\
  Random & 67.72 & 79.34 & 83.19 & 65.85 & 82.79 & \underline{75.78} & \underline{1.08} & 52.25 & 7.00 \\
  \midrule
  DSIR & 67.52 & 77.92 & 83.02 & 65.24 & 81.80 & 75.10 & 1.76 & 50.77 & 5.52 \\
  LESS & 67.45 & 74.47 & 83.62 & 69.51 & 79.15 & 74.84 & 2.02 & 49.19 & 3.94 \\
  TSDS & 68.03 & 78.51 & 82.34 & 63.41 & 83.17 & 75.09 & 1.77 & 50.63 & 5.38 \\
  NICE & 67.01 & 76.12 & 83.19 & 67.68 & 79.83 & 74.77 & 2.09 & 49.58 & 4.33 \\
  \midrule
  GrADS & 67.69 & 78.61 & 84.13 & 64.02 & 83.93 & 75.68 & 1.18 & 51.77 & 6.52 \\
  OGS & 67.60 & 73.64 & 83.45 & 67.68 & 83.85 & 75.24 & 1.62 & 51.42 & 6.17 \\
  \midrule
  Selective DPO & 64.86 & 68.59 & 82.08 & 64.63 & 77.10 & 71.45 & 5.41 & \underline{55.78} & \underline{10.53} \\
  BeeS & 67.54 & 76.49 & 83.53 & 62.80 & 78.92 & 73.86 & 3.00 & 54.62 & 9.37 \\
  PD & 66.86 & 75.30 & 83.28 & 63.41 & 79.15 & 73.60 & 3.26 & 54.97 & 9.72 \\
  \midrule
  BALIGN (Ours) & 67.97 & 78.63 & 82.59 & 65.85 & 84.61 & \textbf{75.93} & \textbf{0.93} & \textbf{56.07} & \textbf{10.82} \\
  \bottomrule
  \end{tabular}
  \vspace{-0.2cm}
  \caption{Detailed general capability and alignment results for harmlessness on the HH-RLHF dataset, using Llama-3.1-8B-Instruct as the base reference model. The best and second-best results are highlighted in bold and underline, respectively, excluding the base model performance.}
  \label{tab:hh-rlhf-harmless-llama}
\end{table*}

\begin{table*}[h]
  \small
  \centering
  \setlength{\tabcolsep}{4.9pt}
  \centering
  \begin{tabular}{l|ccccc|cc|cc}
  \toprule
  Method & MMLU & IFEval & ARC-Challenge & HumanEval & GSM8K & General & Forgetting & Alignment & Gain \\
  \midrule
  Base (Qwen2.5-7B-Instruct) & 73.49 & 76.37 & 89.51 & 84.15 & 84.08 & 81.52 & -- & 44.11 & -- \\
  \midrule
  Full & 73.00 & 73.53 & 88.74 & 84.15 & 80.74 & 80.03 & 1.49 & 52.17 & 8.06 \\
  Random & 73.16 & 76.70 & 88.99 & 82.93 & 82.34 & 80.82 & 0.70 & 47.57 & 3.46 \\
  \midrule
  DSIR & 73.23 & 76.55 & 90.02 & 81.10 & 83.40 & \underline{80.86} & \underline{0.66} & 47.83 & 3.72 \\
  LESS & 73.47 & 75.51 & 89.42 & 81.10 & 80.14 & 79.93 & 1.59 & 48.62 & 4.51 \\
  TSDS & 73.48 & 76.60 & 88.99 & 82.32 & 82.41 & 80.76 & 0.76 & 46.96 & 2.85 \\
  NICE & 73.39 & 74.79 & 89.33 & 79.88 & 79.76 & 79.43 & 2.09 & 48.10 & 3.99 \\
  \midrule
  GrADS & 72.59 & 77.30 & 88.99 & 85.37 & 79.68 & 80.79 & 0.73 & 49.41 & 5.30 \\
  OGS & 72.27 & 74.57 & 89.42 & 83.54 & 77.63 & 79.49 & 2.03 & 48.05 & 3.94 \\
  \midrule
  Selective DPO & 72.33 & 71.53 & 88.74 & 79.27 & 78.62 & 78.10 & 3.42 & \textbf{53.02} & \textbf{8.91} \\
  BeeS & 72.82 & 75.09 & 89.25 & 81.71 & 83.09 & 80.39 & 1.13 & 50.63 & 6.52 \\
  PD & 73.02 & 74.91 & 89.59 & 82.32 & 84.15 & 80.80 & 0.72 & 49.54 & 5.43 \\
  \midrule
  BALIGN (Ours) & 73.05 & 75.34 & 89.33 & 85.37 & 83.85 & \textbf{81.39} & \textbf{0.13} & \underline{52.48} & \underline{8.37} \\
  \bottomrule
  \end{tabular}
  \vspace{-0.2cm}
  \caption{Detailed general capability and alignment results for harmlessness on the HH-RLHF dataset, using Qwen2.5-7B-Instruct as the base reference model. The best and second-best results are highlighted in bold and underline, respectively, excluding the base model performance.}
  \label{tab:hh-rlhf-harmless-qwen}
\end{table*}

\end{document}